\documentclass{article}

\PassOptionsToPackage{numbers, compress}{natbib}

\usepackage[final]{nips_2016}

\usepackage[utf8]{inputenc} \usepackage[T1]{fontenc}    \usepackage{hyperref}       \usepackage{url}            \usepackage{booktabs}       \usepackage{amsfonts}       \usepackage{nicefrac}       \usepackage{microtype}      

\usepackage{graphicx} \usepackage{subfig}
\usepackage{floatrow} \usepackage{wrapfig}

\usepackage{algorithm}
\usepackage{algorithmicx}
\usepackage{algpseudocode}

\usepackage{array}

\usepackage{xspace}

\usepackage{enumitem}
\usepackage{amsmath}
\usepackage{amssymb}
\usepackage{amsthm} 
\usepackage[svgnames]{xcolor}

\newcommand{\rcolb}[1]{\textcolor{red!90}{\textbf{#1}}}

\newcommand{\bcolb}[1]{\textcolor{blue!90}{\textbf{#1}}}

\newlength{\minipagewidth}
\newlength{\minipagewidthx}
\usepackage{calc}

\makeatletter{}\def\:#1{\protect \ifmmode {\mathbf{#1}} \else {\textbf{#1}} \fi}

\newcommand*{\MyDef}{\mathrm{\tiny def}}
\newcommand*{\eqdefU}{\ensuremath{\mathop{\overset{\MyDef}{=}}}}\newcommand*{\eqdef}{\mathop{\overset{\MyDef}{\resizebox{\widthof{\eqdefU}}{\heightof{=}}{=}}}}

\renewcommand{\complement}{\mathsf{c}}
\newcommand{\CommaBin}{\mathbin{\raisebox{0.5ex}{,}}}

\newcommand{\X}{\mathcal{V}}

\newcommand{\indfunc}{\mathbb{I}}

\renewcommand{\Re}{\mathbb{R}}

\newcommand{\wt}[1]{\widetilde{#1}}
\newcommand{\wh}[1]{\widehat{#1}}

\newcommand{\wb}[1]{\overline{#1}}
\newcommand{\transp}{\mathsf{T}}

\DeclareMathOperator*{\Tr}{Tr}

\DeclareMathOperator*{\Ker}{Ker}

\newcommand{\norm}[2]{\left\Vert #1 \right\Vert_{#2}}
\newcommand{\normsmall}[1]{\Vert #1 \Vert}

\newcommand{\probability}{\mathbb{P}}

\DeclareMathOperator*{\expectedvalue}{\mathbb{E}}

\newcommand{\condbar}{\;\middle|\;}

\renewcommand{\Re}{\mathbb{R}}

\newcommand{\Gg}{\mathcal{G}}
\newcommand{\Hg}{\mathcal{H}}

\newcommand{\edgeset}{\mathcal{E}}

\newcommand{\F}{\mathcal{F}}
\newcommand{\Fp}{\mathcal{F}}

\newcommand{\vareps}{\varepsilon}
\renewcommand{\epsilon}{\varepsilon}
\newcommand{\bigotime}{\mathcal{O}}

\newtheorem{theorem}{Theorem}
\newtheorem{lemma}{Lemma}
\newtheorem{definition}{Definition}
\newtheorem{proposition}{Proposition}

\title{Analysis of Kelner and Levin graph sparsification algorithm for a streaming setting}

\author{Daniele Calandriello   \hspace{2em}  Alessandro Lazaric \hspace{2em} Michal Valko\\ SequeL team, INRIA Lille - Nord Europe, France\\ \texttt{\small \{daniele.calandriello, alessandro.lazaric, michal.valko\}@inria.fr}}

\begin{document}

\maketitle

\begin{abstract}
We derive a new proof to show that the incremental resparsification
algorithm proposed by \citet{kelner_spectral_2013} produces a spectral
sparsifier in high probability. We rigorously take into
account the dependencies across subsequent resparsifications using
martingale inequalities, fixing a flaw in the original analysis.
\end{abstract}

\makeatletter{}
\vspace{-0.05in}
\section{Introduction}
\vspace{-0.05in}

\citet{kelner_spectral_2013} introduced a simple single-pass approach to generate a
spectral sparsifier of a graph in the semi-streaming setting, where edges
are received one at a time. They store only an intermediate,
approximate sparsifier and every time a new edge arrives, it is added to it.
Whenever the sparsifier gets too large, they apply a resparsification algorithm
to reduce its size, without compromising its spectral guarantees.

Although the algorithm is intuitive and simple to implement, the original
proof presented in their paper is incomplete,
as originally pointed out in \citet{cohen_online_2016}.
In particular, \citet{kelner_spectral_2013} relies on a concentration
inequality for independent random variables, while in the sparsification algorithm the probability of keeping edge $e$ in the sparsifer at
step $s$ does depend on whether other edges $e'$ have been included in the sparsifier at previous iterations. This structure introduces subtle statistical dependencies through different
iterations of the algorithm, and a more careful analysis is necessary.

In addition to pointing out the problems with the original proof in \cite{kelner_spectral_2013},
\citet{cohen_online_2016} introduces a new algorithm to construct
a sparsifer in a semi-streaming setting but, differently from the original algorithm
in \cite{kelner_spectral_2013}, interactions between iterations are avoided
because the algorithm proposed in \cite{cohen_online_2016}
never drops an edge once it is introduced in the sparsifier.
Another alternative algorithm, this time capable of dropping
included edges, is presented in \citet{pachocki2016analysis} together with a
rigorous proof that takes into account all the dependencies between edges and
between iterations. While the final result in \cite{pachocki2016analysis} guarantees that a valid spectral
sparisfier is generated at each iteration, the proposed algorithm is still
different from the one originally proposed by \citet{kelner_spectral_2013}.

In this note, we derive an alternative proof for the original \citet{kelner_spectral_2013}
algorithm, using arguments similar to \cite{pachocki2016analysis}. In particular,
it is possible to formalize and analyze the edge selection process as a martingale, obtaining
strong concentration guarantees while rigorously taking into account 
the dependencies across the iterations of the algorithm.
 
\makeatletter{}
\vspace{-0.05in}
\section{Background}
\vspace{-0.05in}

\subsection{Notation}
We use lowercase letters $a$ for scalars, bold lowercase letters $\:a$
for vectors and uppercase bold letters $\:A$ for matrices.
We write $\:A \preceq \:B$ for the L{\"o}wner ordering of matrices $\:A$ and $\:B$
when $\:B - \:A$ is positive semi-definite (PSD).

We denote with $\Gg = (\X,\edgeset)$ an undirected weighted graph with $n$
vertices $\X$ and $m$ edges $\edgeset$. Associated with each edge $e_{i,j}\in
\edgeset$ there is a weight $a_{e_{i,j}}$ (shortened $a_{e}$) measuring the ``distance'' between
vertex~$i$ and vertex $j$.\footnote{The graph $\Gg$ can be either constructed
from raw data (e.g., building a $k$-nn graph with an exponential kernel) or it
can be provided directly as input (e.g., in social networks).}
Throughout the rest of the paper, we assume that the weights $a_{e}$ are bounded,
in particular we assume $a_{\max} = \max_{e \in \edgeset} a_{e}$ is smaller
than 1, $a_{\min} = \min_{e \in \edgeset} a_{e}$ is strictly greater than~0,
and that $\kappa^2 = a_{\max}/a_{\min} = \bigotime(\mbox{poly}(n))$,
which is always true for unweighted graphs.
Given two graphs $\Gg$ and $\Gg'$ over the same set of nodes~$\X$, we denote by
$\Gg+\Gg'$ the graph obtained by summing the weights of the edges of $\Gg'$ and
$\Gg$.

Given the
weighted adjacency matrix $\:A_{\Gg}$ and the degree matrix $\:D_{\Gg}$, the
Laplacian of $\Gg$ is the PSD matrix defined as $\:L_\Gg = \:D_\Gg - \:A_\Gg$.
Furthermore, we assume that $\Gg$ is connected and thus has only one eigenvalue
equal to $0$ and $\Ker(\:L_\Gg)=\:1$. Let $\:L_\Gg^+$ be the pseudoinverse of $\:L_\Gg$, and
$\:L_{\Gg}^{-1/2} = (\:L_\Gg^+)^{1/2}$.
For any node $i=1,\ldots,n$, we denote with $\mathbf{\chi}_i\in\Re^n$ the
indicator vector so that $\:b_e = \mathbf{\chi}_i - \mathbf{\chi}_j$ is the ``edge'' vector. If we
denote with $\:B_{\Gg}$ the $m \times n$ signed edge-vertex incidence matrix,
then the Laplacian matrix can be written as $\:L_{\Gg} = \sum_e a_e \:b_e \:b_e^\transp = \:B_\Gg^\transp
\:E_\Gg \:B_\Gg$, where $\:E_\Gg$ is the $m \times m$ diagonal matrix with
$\:E_{\Gg}(e,e) = a_e$.

We indicate with $\:P = \:L_\Gg
\:L_\Gg^+$ the matrix of the orthogonal projection on the $n-1$ dimensional space ortoghonal to
the all one vector $\:1$. Since the Laplacian of any connected graph $\Gg$ has a
null space equal to $\:1$, then $\:P$ is invariant w.r.t. the specific graph
$\Gg$ on $n$ vertices used to defined it.  
 Alternatively, the projection matrix $\:P$ can be obtained as $\:P = 
\:L_\Gg^+\:L_\Gg = \:L_\Gg^{-1/2} \:L_\Gg \:L_\Gg^{-1/2}$.
Finally, let $\:v_{e} = \sqrt{a_{e}}\:L_\Gg^{-1/2} \:b_{e}$, then we have $\:P = \sum_e \:v_e \:v_e^\transp$.

\subsection{Spectral Sparsification in the Semi-Streaming Setting}

A graph $\Hg$ is a spectral sparsifier of $\Gg$ if the whole spectrum of the original graph is well approximated by using only a small portion of its edges. More formally,

\begin{definition}\label{def:eps-sparsifier}
A $ 1 \pm \vareps$ spectral sparsifier of $\Gg$ is a graph $\Hg \subseteq \Gg$ such that for all $\:x\in\Re^n$
\begin{align*}
(1-\vareps)\:x^\transp \:L_\Gg \:x \leq \:x^\transp \:L_\Hg \:x \leq (1 + \vareps) \:x^\transp \:L_\Gg \:x.
\end{align*}
\end{definition}
Spectral sparsifiers store most of the spectral information of the original graph
in a very sparse subgraph. Because of this, they are easy to store in memory
and are used to provide fast approximation to many quantities that are
expensive to compute on the original large graph.

After showing that every graph admits a sparsifier with
$\bigotime(n \log(n)/\varepsilon^2)$ edges,
\citet{spielman2011graph} proposed a sampling algorithm to easily construct
one using the effective resistance of the edges of $\Gg$.

\begin{definition}
The effective resistance of an edge $e$ in graph $\Gg$
is defined as $r_e = \:b_e^\transp \:L_\Gg^+ \:b_e$.
The total weighted sum of effective resistances in a graph is the same for all graphs,
and is equal to
$\sum_e a_e r_e~=~\Tr(\:E_{\Gg} \:B_{\Gg} \:L_\Gg^+ \:B_{\Gg}^\transp) = \Tr(\:L_\Gg \:L_{\Gg}^+) = n-1$.
\end{definition}

Intuitively, the effective resistance encodes the importance of an edge
in preserving the minimum distance between two nodes. If an edge is the only
connection between two parts of the graph, its $r_e$ is large. On the other hand, 
if there are multiple parallel paths across many edges to connect two nodes,
the effective resistance of an edge between the two nodes will be small,
similarly to actual resistances in parallel in
an electrical network. An important consequence of this definition is that
adding edges to a graph can only reduce the effective resistance of other edges,
because it can only introduce new alternative (parallel) paths
in the graph. To prove this formally, consider a graph $\Gg$ and a new
set of edges $\Gamma$. Then we have $\:L_{\Gg} \preceq \:L_{\Gg + \Gamma}$
and therefore 
$\:b_e^\transp \:L_\Gg^+ \:b_e \geq \:b_e^\transp \:L_{\Gg + \Gamma}^+ \:b_e$.

\citet{spielman2011graph} proved that sampling the edges of $\Gg$ with
replacement using a distribution proportional to their effective resistance
produces a spectral sparsifier $\Hg$ of size $\bigotime(n
\log(n)/\varepsilon^2)$ with high probability. The main issue of this approach
is that we want to compute a sparsifier to avoid storing the whole Laplacian,
but we need to store and (pseudo-)invert the Laplacian to compute exact
effective resistances to construct the sparsifier. \citet{spielman2011graph}
showed that this issue can be resolved by computing sufficiently accurate
approximation of the effective resistances.

\begin{definition}
An approximate effective resistance $\wt{r}_e$ is called $\alpha$-accurate for
$\alpha \geq 1$ if it
satisfies
\begin{align*}
    \frac{1}{\alpha} r_e \leq \wt{r}_e \leq \alpha r_e.
\end{align*}
\end{definition}

In particular, \citet[Corollary 6]{spielman2011graph} showed that batch sampling $\bigotime(\alpha^2 n \log(n)/\varepsilon^2)$ edges proportionally
to their $\alpha$-accurate approximate effective resistances is enough to guarantee that the resulting graph is a $1\pm\vareps$-sparsifier.
Building on this result, \citet{kelner_spectral_2013} propose a sequential algorithm (summarized in Alg.~\ref{alg:kl_sequence})
that can emulate the batch sampling of \cite{spielman2011graph} in
a semi-streaming setting and incrementally construct a sparsifier,
without having to fully store and invert the input Laplacian.

\begin{algorithm}[t]
\begin{algorithmic}[1]
    \Require Graph $\Gg$, weights $a_e$ for all edges in $\Gg$.
    \Ensure $\Hg_{\tau}$, a $1 \pm \vareps$ sparsifier of $\Gg$
    \State Set space budget $N = 40\alpha^2 n \log^2 (3\kappa m/\delta) / \vareps^2$
    \State Partition $\Gg$ into $\tau = \lceil m/N \rceil$ blocks $\{\Gamma_{s}\}_{s=0}^\tau$ such that
            $\Gg = \Gg_{\tau} = \sum_{s=0}^{\tau} \Gamma_s$
    \State Initialize $\Hg_{0} = \emptyset$, $\{\wt{p}_{e,0}\} = \emptyset$
    \For{all  $s \in [1,\dots,\tau]$}
        \State Use Algorithm \ref{alg:kl_resparsify} to compute an updated sparsifier $\Hg_s$
        of $\Gg_{s} = \Gg_{s-1} + \Gamma_s$ and new probabilities $\{\wt{p}_{s,e}\}$ for all edges in $\Hg_{s}$.
    \EndFor
    \State Return $\Hg_{\tau}$
\end{algorithmic}
\caption{Kelner and Levin \cite{kelner_spectral_2013} stream sparsification algorithm.}\label{alg:kl_sequence}
\end{algorithm}

\begin{algorithm}[t]
\begin{algorithmic}[1]
    \Require $\Hg_{s-1}$,$\Gamma_s$, the set of previous probabilities $\{\wt{p}_{s-1,e}\}$
        for all edges in $\Hg_{s-1}$ and the weights $a_e$ for all edges in $\Hg_{s-1} + \Gamma_s$
    \Ensure $\Hg_{s}$, a $1 \pm \vareps$ sparsifier of $\Gg_{s} = \Gg_{s-1} + \Gamma_{s}$ and new 
    probabilities $\wt{p}_{s,e}$.
        \State Compute estimates of all $\wt{r}_{s,e}$ in $\Hg_{s-1}+\Gamma_{s}$ using a fast SDD solver (\cite[Theorem~3]{kelner_spectral_2013})
    \State Compute new probabilities $\wt{p}_{s,e}~=~(a_{e}\wt{r}_{s,e})/(\alpha(n-1))$
    \For{all edges $e \in \Hg_{s-1}$}
        \State $\wt{p}_{s,e} \leftarrow \min\{\wt{p}_{s-1,e},\wt{p}_{s,e}\}$ \label{alg:tag-take-minimum}
    \EndFor
    \State Initialize $\Hg_{s} = \emptyset$
    \For{all edges $e \in \Hg_{s-1}$}
    \State With probability $\wt{p}_{s,e}/\wt{p}_{s-1,e}$ add edge $e$ to $\Hg_{s}$ with weight $a_e/(N \wt{p}_{s,e})$ \label{alg:tag-probability-ratio}
    \EndFor
    \For{all edges $e \in \Gamma_s$}
    \For{$ i = 1 $ to $N$}
    \State With probability $\wt{p}_{s,e}$ add a copy of edge $e$ to $\Hg_s$ with weight $a_e/(N \wt{p}_{s,e})$
    \EndFor
    \EndFor
\end{algorithmic}
\caption{Kelner and Levin \cite{kelner_spectral_2013} resparsification algorithm at step $s$.}\label{alg:kl_resparsify}
\end{algorithm}

In a semi-streaming setting the graph $\Gg$ is split int $\tau = \lceil m/N \rceil$ blocks $\Gamma_s$, with $s \in [1,2,\dots,\tau]$, where each block
is a subset of $N$ edges
such that $\Gg = \Gg_\tau = \Gg_{\tau-1} + \Gamma_\tau = \Gamma_1 + \Gamma_2 + \dots + \Gamma_\tau$.
Associated with each of the $\Gg_s$ partial graph, we can define its respective
effective resistances $r_{s,e}$, and the sampling probabilities $p_{s,e} = (a_{e}r_{s,e})/(n-1)$.
    Starting ($s=0$) from an empty sparsifier $\Hg_0 = \emptyset$, the algorithm
alternates between two phases. In the first phase, the algorithm reads $N$ edges from
the stream to build a new
block $\Gamma_s$, and it combines it with the previous sparsifier $\Hg_{s-1}$
to construct $\Hg_{s-1} + \Gamma_s$. This phase sees an increase in memory usage,
because the algorithm needs to store the newly arrived edges in addition to the
sparsifier. In the second phase the graph $\Hg_{s-1} + \Gamma_s$ is used
together with a fast SDD solver to compute
$\alpha$-accurate estimates $\wt{r}_{s,e}$ of the effective resistance of all the edges in it.
These approximate effective resistances $\wt{r}_{s,e}$
are used to compute approximate probabilities $\wt{p}_{s,e}$,
and according to these approximate probabilities
each of the edges in $\Hg_{s-1}$ and $\Gamma_s$ is added to the new sparsifier
$\Hg_{s}$ or discarded forever freeing up memory.
Choosing carefully the size of the blocks $\Gamma_s$ to be close to the size of the
sparsifiers $\Hg_s$ allows the algorithm to run efficiently in a small fixed
space and produce a valid sparsifier at the end of each
iteration.\footnote{Throughout this note, we consider that the decomposition of
$\Gg$ into blocks is such that all intermediate graphs $\Gg_s$ are fully
connected. Whenever this is not the case, the algorithm should be adjusted to
run separately on all the components of the graph.}
For more details on the time complexity analysis and the implementation details
on how to obtain $\alpha$-approximate effective resistances using
a valid sparsifier and a fast SDD solver we refer to the original paper
\cite{kelner_spectral_2013}.

The main result of \citet{kelner_spectral_2013} is the following theorem.

\begin{theorem}\label{thm:main-theorem}
    Let $\Hg_s$ be the sparsifier returned by Algorithm \ref{alg:kl_sequence}
    after $s$ resparsifications (after streaming the first $s$ blocks).
    If $\alpha \leq \sqrt{\kappa n/3}$ and $\kappa^2 = \bigotime(\text{poly}(n))$,
    with probability $1-\delta$ all sparsifiers $\Hg_s$ from the beginning
    of the algorithm to its end ($\forall s \in [1,\dots,\tau]$)
    are valid $(1 \pm \vareps)$ spectral sparsifiers of their corresponding
    partial graph $\Gg_s$ and the number of edges in each of the $\Hg_s$
    sparsifiers is $\bigotime(n \log^2(\kappa m)/\vareps^2) = \bigotime(n \log^2(n)/\vareps^2) $.
\end{theorem}

At the core of the original proof of this theorem, Kelner and Levin rely on a concentration inequality for independent random variables in~\cite{vershynin_note_2009}. Unfortunately, it is not possible to directly use this result since the probability that an edge $e$ is included in $\Hg_s$ does indeed depend on all the edges that were included in $\Hg_{s-1}$ through the computation of $\wt{r}_{s,e}$. The sparsifier $\Hg_{s-1}$ in turn is generated from $\Hg_{s-2}$ and so on. As a result, the probability that an edge $e$ is present in the final graph $\Hg_\tau$ is strictly dependent on the other edges.
In the following we rigorously take into account the interactions across
iterations and provide a new proof for Theorem~\ref{thm:main-theorem}
which confirms its original statement, thus proving the correctness of \cite{kelner_spectral_2013}'s algorithm.

\makeatletter{}
\vspace{-0.05in}
\section{Proof}
\vspace{-0.05in}

\begin{algorithm}[t]
\begin{algorithmic}[1]
    \Require Graph $\Gg$, weights $a_e$ for all edges in $\Gg$.
    \Ensure $\Hg_{m}$, a $1 \pm \vareps$ sparsifier of $\Gg$
    \State Set space budget $N = 40\alpha^2 n \log^2 (3\kappa m/\delta) / \vareps^2$
    \State Set $\wh{z}_{s,e,j} = 1$, $\wt{p}_{s,e} = 1$ for all $e=1,\ldots,m$, $s=0,\ldots,m$, $j=1,\ldots,N$
    \For{$s=1,\ldots,m$} \Comment{Stream over all edges in $\Gg$} \label{alg:tag-outer-for-loop}
        \For{$e=1,\ldots,m$}
            \If{$e \leq s$}
            \State Compute $\wt{r}_{s,e}$ using $\Hg_{s-1}$ and fast SDD solver
            \State Compute $\wt{p}_{s,e} = \min\left\{\frac{a_{e}\wt{r}_{s,e}}{\alpha(n-1)} , \wt{p}_{s-1,e} \right\}$
            \EndIf
            \For{$j=1,\ldots,N$} \Comment{Loop over $N$ trials (i.e., copies)}
            \If{$\wh{z}_{s-1,e,j} = 1$}
            \State Sample $\wh{z}_{s,e,j} \sim \mathcal{B}\Big( \frac{\wt{p}_{s,e}}{\wt{p}_{s-1,e}} \Big)$\label{alg:tag-keep-edge}
            \Else
            \State $\wh{z}_{s,e,j} = 0$
            \EndIf
            \EndFor
        \EndFor
        \State Set $\Hg_s = \emptyset$
        \For{$e=1,\ldots,s$, $j=1,\ldots,N$} \Comment{Construction of the new sparsifier}
            \If{$\wh{z}_{s,e,j} = 1$}
	        \State Add edge $e$ to $\Hg_s$ with weight $a_e/(N \wt{p}_{s,e})$
           \EndIf
        \EndFor
    \EndFor
\end{algorithmic}
\caption{Equivalent formulation of Kelner and Levin's algorithm.}\label{alg:kl_theory}
\end{algorithm}

\textbf{Step 1 (the \textit{theoretical} algorithm).} 
In order to simplify the analysis, we introduce an equivalent formulation of Alg.~\ref{alg:kl_sequence}.
In Alg.~\ref{alg:kl_theory} we consider the case where the blocks $\Gamma_s$
contain only a single edge and the algorithm performs $m$ resparsifications
over the course of the whole stream of $m$ edges (loop at line 3). This is a
wasteful approach and more practical methods
(such as Alg.~\ref{alg:kl_sequence})
choose to resparsify only when $\Gamma_s$
is larger than $\Hg_{s-1}$ in order to save time without increasing
the asymptotic space complexity of the algorithm. Nonetheless, the single edge setting can be used to emulate
any larger choice for the block size $\Gamma_s$, and therefore we only need
to prove our result in this setting for it to hold in any other case.

In Alg.~\ref{alg:kl_theory}, we denote by $\wh{z}_{s,e,j}$ the Bernoulli random variable 
that indicates whether copy $j$ of edge~$e$ is present in the sparsifier $\Hg_s$ at step $s$. While an edge $e$ can be present in $\Hg_s$ only if $e\leq s$, we initialize $\wh{z}_{s,e,j}=1$ for all $e>s$ for notational convenience. 
The way these variables are generated is equivalent to lines 7-14 in Alg.~\ref{alg:kl_resparsify} so that at iteration $s$, each copy of an edge $e$ already in the sparsifier $\Hg_{s-1}$ is kept with probability $\wt{p}_{s,e}/\wt{p}_{s-1,e}$, while a copy of the new edge $s$ is added with probability $\wt{p}_{s,e}$. For any edge $e>s$ (i.e., not processed yet in Alg.~\ref{alg:kl_resparsify}) we initialize $\wt{p}_{s,e} = 1$ and thus the sampling in line 10 of Alg.~\ref{alg:kl_theory} always returns $\wh{z}_{s,e,j}=1$. Since edges are added with weights $a_e/(N \wt{p}_{s,e})$, after processing $s \leq m$ edges, the Laplacian of the sparsifier $\Hg_s$ can be written as
\begin{align*}
\:L_{\Hg_{s}} = \sum_{e=1}^s \sum_{j=1}^N \frac{a_e}{N\wt{p}_{s,e}}\wh{z}_{s,e,j}\:b_e\:b_e^\transp.
\end{align*}

\textbf{Step 2 (filtration).}
A convenient way to treat the indicator variables $\wh{z}_{s,e,j}$ is to define them recursively as
\begin{align*}
    \wh{z}_{s,e,j} \eqdef \indfunc\left\{ u_{s,e,j} \leq \frac{\wt{p}_{s,e}}{\wt{p}_{s-1,e}}\right\} \wh{z}_{s-1,e,j},
\end{align*}
where $u_{s,e,j} \sim \mathcal{U}(0,1)$ is a uniform random variable used
to compute the $\wt{p}_{s,e}/\wt{p}_{s-1,e}$ coin flip,
and~$\wt{p}_{s,e}$ are the approximate probabilities computed at step $s$
according to the definition in Algorithm~\ref{alg:kl_resparsify},
using the SDD solver, the sparsifier $\Hg_{s-1}$ and the new edges $\Gamma_{s}$.
This formulation allows us to define the stochastic properties of variables $\wh{z}_{s,e,j}$ in a convenient way. We first arrange the indices $s$, $e$, and $j$ into a linear index $r=\{s,e,j\}$ in the range $[1,\dots,m^2N]$,
obtained as $r = \{s,e,j\} = (s-1)mN + (e-1)N + j$. Following the structure of Alg.~\ref{alg:kl_theory}, the linearization wraps first when $j$ hits its limit, and then when $e$ and finally
$s$ do the same, such that for any $s<m$, $e<m$, and $j<N$, we have
\begin{align*}
    &\{s,e,j\}+1 = \{s,e,j+1\},\\
    & \{s,e,N\} + 1 = \{s,e+1,1\},\\
    & \{s,m,N\} + 1 = \{s+1,1,1\}.
\end{align*}
It is easy to see that the checkpoints $\{s,m,N\}$ correspond to a full
iteration (Alg. \ref{alg:kl_theory}, line \ref{alg:tag-outer-for-loop}) of the
algorithm. Let $\F_{\{s,e,j\}}$ be the filtration containing all the
realizations of the uniform random variables $u_{s,e,j}$ up to the step
$\{s,e,j\}$, that is $\F_{\{s,e,j\}} = \{ u_{s',e',j'}, \forall\{s',e',j'\}
\leq \{s,e,j\}\}$. Again, we notice that $\F_{\{s,m,N\}}$ defines the state of
the algorithm after completing iteration $s$. Since $\wt{r}_{s,e}$ and
$\wt{p}_{s,e}$ are computed at the beginning of iteration $s$ using the
sparisfier $\Hg_{s-1}$, they are fully determined by $\F_{\{s-1,m,N\}}$.
Furthermore, since $\F_{\{s-1,m,N\}}$ also defines the values of all indicator
variables~$\wh{z}_{s',e,j}$ up to $\wh{z}_{s-1,e,j}$ for any $e$ and $j$, we have that
all the Bernoulli variables $\wh{z}_{s,e,j}$ at iteration $s$ are conditionally
independent  given $\F_{\{s-1,m,N\}}$. In other words, we have that for any
$e'$, and $j'$ such that $\{s,1,1\} \leq \{s,e',j'\} <\{s,e,j\}$ the following
random variables are equal in distribution
\begin{align}\label{eq:distro.z}
\wh{z}_{s,e,j} \big| \F_{\{s,e',j'\}} = \wh{z}_{s,e,j} \big| \F_{\{s-1,m, N\}} \sim \mathcal{B}\Big( \frac{\wt{p}_{s,e}}{\wt{p}_{s-1,e}} \Big)
\end{align}
and for any $e'$, and $j'$ such that $\{s,1,1\} \leq \{s,e',j'\} \leq \{s,m,N\} $ and $\{s,e,j\} \neq \{s,e',j'\}$
\begin{align}\label{eq:distro.z.indep}
\wh{z}_{s,e,j} \big| \F_{\{s-1,m, N\}} \perp \wh{z}_{s,e',j'} \big| \F_{\{s-1,m, N\}}.
\end{align}

\textbf{Step 3 (the projection error).} While our objective is to show that $\Hg_m$ is a $1\pm\vareps$-sparsifier, following~\cite{kelner_spectral_2013} we study the related objective of defining an approximate projection matrix that is close in $\ell_2$-norm to the original projection matrix $\:P$. In fact, the two objectives are strictly related as shown in the following proposition.

\begin{proposition}[\cite{kelner_spectral_2013}]\label{prop:sparsifier-identity}
    Given $\:v_{e} = \sqrt{a_{e}}\:L_\Gg^{-1/2} \:b_{e}$, let $\mathcal{I} = [e_1, e_2, \dots, e_N] \in \edgeset^N$ be a subset of edges of $\Gg$ and 
    $\wt{\:P} = \sum_{i=1}^{N} w_i \:v_{e_i} \:v_{e_i}^\transp$ an approximate projection matrix with weights $\{w_i\}_{i=1}^N$. If the weights are such that
    \begin{align}\label{eq:approx.projection}
         \norm{\:P - \sum_{i=1}^{N} w_i \:v_{e_i} \:v_{e_i}^\transp}{2} = \norm{\:P - \wt{\:P}}{2} \leq \vareps,
    \end{align}
then the graph $\Hg$ obtained by adding edges
    $e_i\in\mathcal{I}$ with weights $a_{e_i}w_i$ is a $(1\pm\vareps)$-sparsifier of $\Gg$.
\end{proposition}

Using the notation of Alg.~\ref{alg:kl_theory}, the approximate projection matrix is defined as
\begin{align*}
\wt{\:P} \eqdef \frac{1}{N} \sum_{j=1}^N \sum_{e=1}^m\frac{\wh{z}_{m,e,j}}{\wt{p}_{m,e}}\:v_e\:v_e^\transp,
\end{align*}
and thus the previous proposition suggests that to prove that the graph $\Hg_m$ returned by Algorithm $\ref{alg:kl_sequence}$
after $m$ steps is a sparsifier, it is sufficient to show that
\begin{align*}
         \norm{\:P - \wt{\:P}}{2}
=\norm{\frac{1}{N}\sum_{j=1}^{N}\sum_{e=1}^m\left(1-\frac{\wh{z}_{m,e,j}}{\wt{p}_{m,e}}\right)\:v_{e}\:v_{e}^\transp}{2} \leq \epsilon.
\end{align*}
In order to study this quantity, we need to analyze how the projection error evolves over iterations. To this end, we introduce term $\wh{\:Y}_{\{s,e,j\}}$
which denotes the projection error at the end of step $\{s,e,j\}$ of Algorithm~\ref{alg:kl_theory}.
\begin{align*}
    \wh{\:Y}_{\{s,e,j\}} \eqdef &
     \underbrace{\frac{1}{N}\sum_{k=1}^{e-1}\sum_{l=1}^{N}\left(1-\frac{\wh{z}_{s,k,l}}{\wt{p}_{s,k}}\right)\:v_{k}\:v_{k}^\transp}_{\text{edges already processed at step $s$}}\\
    &+\underbrace{\frac{1}{N}\left(\sum_{l=1}^{j}\left(1-\frac{\wh{z}_{s,e,l}}{\wt{p}_{s,e}}\right) + \sum_{l=j+1}^{N}\left(1-\frac{\wh{z}_{s-1,e,l}}{\wt{p}_{s-1,e}}\right)\right)\:v_{e}\:v_{e}^\transp}_{\text{copy $j$ of edge $e$ getting processed at step $s$ (Alg.~\ref{alg:kl_theory}, line \ref{alg:tag-keep-edge})}}\\
    &+\underbrace{\frac{1}{N}\sum_{k=e+1}^{m}\sum_{l=1}^{N}\left(1-\frac{\wh{z}_{s-1,k,l}}{\wt{p}_{s-1,k}}\right)\:v_{k}\:v_{k}^\transp}_{\text{edges still not processed at step $s$}}\,
\end{align*}
Notice that setting $\wh{z}_{s,e,j}=1$ and $\wt{p}_{s,e}=1$ for any $e>s$ implies that the edges that have not been processed yet do not contribute to the projection error. Finally, notice that at the end of the algorithm we have $\wh{\:Y}_{\{m,m,N\}} = \:P - \wt{\:P}$, which quantifies the error of the output of Algorithm \ref{alg:kl_sequence}.

We are now ready to restate Theorem~\ref{thm:main-theorem} in a more convenient way as
\begin{align*}
    \probability\bigg(\exists  s\in\{1,\ldots,m\}: \underbrace{\vphantom{\sum_{e=1}^m}\normsmall{\wh{\:Y}_{\{s,m,N\}}} \geq \varepsilon}_{A_s} \cup \underbrace{\sum_{e=1}^s \sum_{j=1}^N \wh{z}_{s,e,j} \geq 3N}_{B_s}\bigg) \leq \delta,
\end{align*}
where the first event $A_s$ refers to the case when for any $s \in [1,\dots,m]$ the intermediate graph $\Hg_s$ fails to be a valid sparsifier and the second event $B_s$ considers the event when the memory requirement is not met (i.e., too many edges are kept in the sparsifier $\Hg_s$).

To prove the statement, we decompose the probability of failure as follows.
\begin{align}\label{eq:thm.decomposition}
    &\probability\bigg(\exists  s\in\{1,\ldots,m\}: \normsmall{\wh{\:Y}_{\{s,m,N\}}} \geq \varepsilon \cup \sum_{e=1}^s \sum_{j=1}^N \wh{z}_{s,e,j} \geq 3N\bigg)
    =\probability\bigg(\bigcup_{s = 1}^{m} A_s \cup B_s\bigg)\nonumber\\
    &=\probability\left(\left\{\bigcup_{s = 1}^{m} A_s\right\} \cup \left\{ \bigcup_{s = 1}^{m} B_s\right\}\right)\nonumber\\
    &=\probability\left(\left\{\bigcup_{s = 1}^{m} A_s\right\} \right)
    +\probability\left( \left\{ \bigcup_{s = 1}^{m} B_s\right\}\right)
    -\probability\left(\left\{\bigcup_{s = 1}^{m} A_s\right\} \cap \left\{ \bigcup_{s = 1}^{m} B_s\right\}\right) \nonumber\\
    &=\probability\left(\left\{\bigcup_{s = 1}^{m} A_s\right\} \right)
    +\probability\left(\left\{\bigcup_{s = 1}^{m} B_s\right\} \cap \left\{ \bigcup_{s = 1}^{m} A_s\right\}^{\complement}\right) \nonumber\\
    &=\probability\left(\left\{\bigcup_{s = 1}^{m} A_s\right\} \right)
    +\probability\left(\left\{\bigcup_{s = 1}^{m} B_s\right\} \cap \left\{ \bigcap_{s = 1}^{m} A_s^{\complement}\right\}\right) \nonumber\\
    &=\probability\left(\left\{\bigcup_{s = 1}^{m} A_s\right\} \right)
    +\probability\left(\bigcup_{s = 1}^{m} \left\{B_s \cap \left\{ \bigcap_{s' = 1}^{m} A_{s'}^{\complement}\right\}\right\}\right)\nonumber
\end{align}
Taking the last formulation and replacing the definitions of $A_s$ and $B_s$, we get
\begin{align}
    &\probability\bigg(\exists  s\in\{1,\ldots,m\}: \normsmall{\wh{\:Y}_{\{s,m,N\}}} \geq \varepsilon \cup \sum_{e=1}^s \sum_{j=1}^N \wh{z}_{s,e,j} \geq 3N\bigg)\nonumber\\
    &=\probability\left(\left\{\bigcup_{s = 1}^{m} \normsmall{\wh{\:Y}_{\{s,m,N\}}} \geq \varepsilon \right\} \right)
    +\probability\left(\bigcup_{s = 1}^{m} \left\{\sum_{e=1}^s \sum_{j=1}^N \wh{z}_{s,e,j} \geq 3N \cap \left\{ \bigcap_{s' = 1}^{m} \normsmall{\wh{\:Y}_{\{s',m,N\}}} \leq \varepsilon\right\}\right\}\right)\nonumber\\
    &=\probability\left(\exists  s \in \{1, \dots, m\}:  \normsmall{\wh{\:Y}_{\{s,m,N\}}} \geq \varepsilon  \right)\nonumber\\
    &\quad +\probability\left(\exists  s\in \{1, \dots, m\}: \sum_{e=1}^s \sum_{j=1}^N \wh{z}_{s,e,j} \geq 3N \cap  \left\{\forall  s'\in \{1, \dots, m\}:\normsmall{\wh{\:Y}_{\{s',m,N\}}} \leq \varepsilon\right\}\right)\nonumber\\
    &\leq \sum_{s = 1}^{m} \probability\left(\normsmall{\wh{\:Y}_{\{s,m,N\}}} \geq \varepsilon \right)\nonumber\\
    &\quad +\sum_{s = 1}^{m}\probability\left(\sum_{e=1}^s \sum_{j=1}^N \wh{z}_{s,e,j} \geq 3N \cap \left\{\forall  s' \in \{1, \dots, m\} :  \normsmall{\wh{\:Y}_{\{s',m,N\}}} \leq \varepsilon\right\}\right)\nonumber\\
    &\leq \sum_{s = 1}^{m} \probability\left(\normsmall{\wh{\:Y}_{\{s,m,N\}}} \geq \varepsilon \right)\nonumber\\
    &\quad +\sum_{s = 1}^{m}\probability\left(\sum_{e=1}^s \sum_{j=1}^N \wh{z}_{s,e,j} \geq 3N \cap \left\{\forall  s' \in \{1, \dots, s\} :  \normsmall{\wh{\:Y}_{\{s',m,N\}}} \leq \varepsilon\right\}\right)
\end{align}

\textbf{Step 4 (putting everything together).} In the following sections, we prove the two main lemmas of this note, where we bound the probability of returning a non-spectral sparsifier and the probability of exceeding too much the budget limit $N$. In particular, we derive the two following results.

\begin{lemma}\label{lem:approx-bound}
    \begin{align*}
        \probability\left( \normsmall{\wh{\:Y}_{\{t,m,N\}}} \geq \epsilon\right)
        \leq \frac{\delta}{2m}
    \end{align*}
\end{lemma}

\begin{lemma}\label{lem:space-concentration}
    \begin{align*}
\probability\left(\sum_{e=1}^t \sum_{j=1}^N \wh{z}_{t,e,j} \geq 3N \cap \left\{\forall \; s \in \{1, \dots, t\} :  \normsmall{\wh{\:Y}_{\{s,m,N\}}} \leq \varepsilon\right\}\right)
    \leq \frac{\delta}{2m}
    \end{align*}
\end{lemma}

Combining the two lemmas into Eq.~\ref{eq:thm.decomposition}, we prove
Thm~.\ref{thm:main-theorem} for an algorithm that resparsifies every time a new edge
arrives ($|\Gamma_s| = 1$).
Extending the proof to the case when multiple edges are stored in
$\Gamma_s$ before a new resparsification happens is straightforward.
In the proofs of Lemma \ref{lem:approx-bound} and \ref{lem:space-concentration}
the fact that an edge is unseen (not streamed yet)
is represented by deterministically setting its $\wt{p}_{s,e}$
to 1, while the estimates for seen edges are computed based on the graph.
To represent the arrival of an edge $e$ at time $t$
we simply start updating its $\wt{p}_{t,e}$. To take into account $\tau$
resparsifications of large blocks instead of $m$ resparsifications
of single-edge blocks it is sufficient to start updating multiple
$\wt{p}_{t,e}$ at the same step. The rest of the analysis remains unchanged.
 
\makeatletter{}
\vspace{-0.05in}
\section{Proof of Lemma \ref{lem:approx-bound} (bounding $\wh{\bf Y}_{\{s,m,N\}}$)}
\vspace{-0.05in}

\textbf{Step 1 (freezing process).}
We first restate a proposition on the accuracy of the effective resistance estimates.

\begin{proposition}\label{prop:alpha.good}
At iteration $s$ the approximated effective resistance $\wt{r}_{s,e}$ of an edge $e$ in $\Hg_{s-1}+\Gamma_s$ is computed using $\Hg_{s-1} + \Gamma_s$ and the SDD solver. If $\Hg_{s-1}$ is a valid $1\pm\varepsilon$-sparsifier of $\Gg_{s-1}$, then $\wt{r}_{s,e}$ is $\alpha$-accurate.
\end{proposition}

Given $\alpha$-accurate effective resistances, the approximate probabilities $\wt{p}_{s,e}$ are defined as
\begin{align*}
    \wt{p}_{s,e} = \min\left\{\frac{a_{e}\wt{r}_{s,e}}{\alpha(n-1)} , \wt{p}_{s-1,e} \right\}.
\end{align*}
As pointed out in Proposition~\ref{prop:alpha.good}, the main issue
is that whenever $\Hg_{s-1}$ is not a valid sparsifier of $\Gg_{s-1}$, the
approximate probabilities $\wt{p}_{s,e}$ returned by the fast SDD solver are
not guaranteed to be $\alpha$-accurate approximations of the true probabilities
$p_{s,e}$. While the overall algorithm may fail in generating a valid sparsifier at some intermediate iteration and yet return a valid sparsifier at the end, we consider an alternative (more pessimistic) process which is ``frozen'' as soon as it constructs an invalid sparsifier.
Consider an alternative process $\:Y_{\{s,e,j\}}$
based on the following definition of approximate probabilities 
\begin{align*}
\wb{p}_{s,e} =\wt{p}_{s,e}
    \indfunc\left\{\normsmall{\:Y_{\{s-1,m,N\}}} \leq \varepsilon\right\}
    +\wb{p}_{s-1,e}\indfunc\left\{\normsmall{\:Y_{\{s-1,m,N\}}} \geq \varepsilon\right\},
\end{align*}
where by Proposition~\ref{prop:sparsifier-identity}, the condition $\normsmall{\:Y_{\{s-1,m,N\}}} \leq \varepsilon$ is equivalent to requiring that $\Hg_{s-1}$ is a valid sparsifier.
This new formulation represents a variant of our algorithm that can detect if the
previous iteration failed to construct a graph that is guaranteed to be a
sparsifier. When this failure happens, the whole process is frozen
and continues until the end without updating anything.
Then we redefine the indicator variable $z_{s,e,j}$ dependent on $\wb{p}_{s,e}$ as
\begin{align*}
    z_{s,e,j} = \indfunc\left\{ u_{s,e,j} \leq \frac{\wb{p}_{s,e}}{\wb{p}_{s-1,e}}\right\} z_{s-1,e,j},
\end{align*}
and then the projection  error process based on them becomes
\begin{align*}
    \:Y_{\{s,e,j\}} &=
    \frac{1}{N}\sum_{k=1}^{e-1}\sum_{l=1}^{N}\left(1-\frac{z_{s,k,l}}{\wb{p}_{s,k}}\right)\:v_{k}\:v_{k}^\transp\\
    &+\frac{1}{N}\left(\sum_{l=1}^{j}\left(1-\frac{z_{s,e,l}}{\wb{p}_{s,e}}\right) + \sum_{l=j+1}^{N}\left(1-\frac{z_{s-1,e,l}}{\wb{p}_{s-1,e}}\right)\right)\:v_{e}\:v_{e}^\transp\\
    &+\frac{1}{N}\sum_{k=e+1}^{m}\sum_{l=1}^{N}\left(1-\frac{z_{s-1,k,l}}{\wb{p}_{s-1,k}}\right)\:v_{k}\:v_{k}^\transp.
\end{align*}
We can see that whenever $\normsmall{\:Y_{\{s,m,N\}}} \geq \varepsilon$ at step $s$,
for all successive steps $s'$ we have $z_{s',e,j}~=~z_{s,e,j}$, or in other words
we never drop or add a new edge and never change their weights, since $\wb{p}_{s,e}$ is constant.
Consequently, if any of the
intermediate elements of the sequence violates the condition $\normsmall{\:Y_{\{s,e,j\}}}~\leq~\varepsilon$, the last element
will violate it too.
For the rest, the sequence behaves exactly
like $\wh{\:Y}_{\{s,e,j\}}$.
Therefore,
\begin{align*}
&\probability\left( \normsmall{\wh{\:Y}_{\{t,m,N\}}} \geq \varepsilon\right)
\leq \probability\Big( \normsmall{\:Y_{\{t,m,N\}}} \geq \varepsilon\Big).
\end{align*}

\textbf{Step 2 (martingale process).} We now proceed by studying the process $\{\:Y_{\{s,e,j\}}\}$ and showing that it is a bounded martingale.
The sequence difference process $\{ \:X_{\{s,e,j\}} \}$ is defined as
$\:X_{\{s,e,j\}}~=~\:Y_{\{s,e,j\}}~-~\:Y_{\{s,e,j\}-1}$, that is
\begin{align*}
    \:X_{\{s,e,j\}}
     &= \frac{1}{N}\left(\frac{z_{s-1,e,j}}{\wb{p}_{s-1,e}} - \frac{z_{s,e,j}}{\wb{p}_{s,e}}\right)\:v_{e}\:v_{e}^\transp.
\end{align*}
In order to show that $\:Y_{\{s,e,j\}}$ is a martingale, it is sufficient to verify the following (equivalent) conditions
\begin{align*}
    \expectedvalue\left[\:Y_{\{s,e,j\}} \condbar \F_{\{s,e,j\}-1}\right]
    =\:Y_{\{s,e,j\}-1} \enspace \Leftrightarrow \enspace
    \expectedvalue\left[\:X_{\{s,e,j\}} \condbar \F_{\{s,e,j\}-1}\right] = \:0.
\end{align*}
We begin by inspecting the conditional random variable $\:X_{\{s,e,j\}} | \F_{\{s,e,j\}-1}$. Given the definition of $\:X_{\{s,e,j\}}$, the conditioning on $\F_{\{s,e,j\}-1}$ determines the values of $z_{s-1,e,j}$ and the approximate probabilities $\wb{p}_{s-1,e}$ and $\wb{p}_{s,e}$. In fact, remember that these quantities are fully determined by the realizations in $\F_{\{s-1,m,N\}}$ which are contained in $\F_{\{s,e,j\}-1}$. As a result, the only stochastic quantity in $\:X_{\{s,e,j\}}$ is the variable $z_{s,e,j}$. Specifically, if $\normsmall{\:Y_{\{s-1,m,N\}}} \geq \varepsilon$,
then we have $\wb{p}_{s,e} = \wb{p}_{s-1,e}$ and $z_{s,e,j} = z_{s-1,e,j}$
(the process is stopped), and the martingale requirement
$ \expectedvalue\left[\:X_{\{s,e,j\}} \condbar \F_{\{s,e,j\}-1}\right] = \:0$
is trivially satisfied.
On the other hand, if $\normsmall{\:Y_{\{s-1,m,N\}}} \leq \varepsilon$ we have,
\begin{align*}
\expectedvalue_{u_{s,e,j}}&\left[\frac{1}{N}
    \left(\frac{z_{s-1,e,j}}{\wb{p}_{s-1,e}} - \frac{z_{s,e,j}}{\wb{p}_{s,e}}\right)\:v_{e}\:v_{e}^\transp \condbar \F_{\{s,e,j\}-1}\right]\\
 &= \frac{1}{N}
    \left( \frac{z_{s-1,e,j}}{\wb{p}_{s-1,e}} -\frac{z_{s-1,e,j}}{\wb{p}_{s,e}}\expectedvalue\left[\indfunc\left\{ u_{s,e,j} \leq \frac{\wb{p}_{s,e}}{\wb{p}_{s-1,e}}\right\}\condbar\F_{\{s,e,j\}-1}\right]\right)\:v_{e}\:v_{e}^\transp\\
&= \frac{1}{N}
    \left(\frac{z_{s-1,e,j}}{\wb{p}_{s-1,e}} - \frac{z_{s-1,e,j}}{\wb{p}_{s,e}}\frac{\wb{p}_{s,e}}{\wb{p}_{s-1,e}}
\right)\:v_{e}\:v_{e}^\transp
= \:0,
\end{align*}
where we use the recursive definition of $z_{s,e,j}$ and the fact that $u_{s,e,j}$ is a uniform random variable in $[0,1]$. This proves that $\:Y_{\{s,e,j\}}$ is indeed a martingale.
We now compute an upper-bound $R$ on the norm of the values of the difference process as
\begin{align*}
&\normsmall{\:X_{\{s,e,j\}}}
= \frac{1}{N} \left|\left(\frac{z_{s-1,e,j}}{\wb{p}_{s-1,e}} - \frac{z_{s,e,j}}{\wb{p}_{s,e}}\right)\right|\normsmall{\:v_{e}\:v_{e}^\transp}
= \frac{1}{N} \left|\left(\frac{z_{s-1,e,j}}{\wb{p}_{s-1,e}} - \frac{z_{s,e,j}}{\wb{p}_{s,e}}\right)\right|\normsmall{\:v_{e}}^2
\\
&\leq \frac{1}{N} \frac{a_e r_{m,e}}{\wb{p}_{s,e}}
\leq \frac{1}{N} \frac{\alpha^2 a_{e} r_{m,e}}{p_{s,e}}
\leq \frac{1}{N} \frac{\alpha^2 a_{e} r_{m,e}}{p_{m,e}}
=\frac{1}{N} \frac{\alpha^2 (n-1)a_e r_{m,e}}{a_e r_{m,e}}
=\frac{\alpha^2 (n-1)}{N} \eqdef R,
\end{align*}
where we use the fact that if, at step $s$, $\normsmall{\:Y_{\{s-1,m,N\}}} \leq \varepsilon$,
then the approximate $\wt{r}_{s,e}$ are
$\alpha$-accurate by Proposition~\ref{prop:alpha.good} and thus 
by definition of $p_{s,e}$,
\begin{align*}
    \frac{p_{s,e}}{\alpha^2} \leq \wb{p}_{s,e} \leq p_{s,e}.
\end{align*}
If instead, $\normsmall{\:Y_{\{s-1,m,N\}}} \geq \varepsilon$,
 the process is stopped and 
$\normsmall{\:X_{\{s,e,j\}}} = \normsmall{\:0} = 0 \leq R$.

\textbf{Step 3 (martingale concentration inequality).}
We are now ready to use a Freedman matrix inequality from \cite{tropp2011freedman} to bound the norm of $\:Y$.

\begin{proposition}[Theorem~1.2~\cite{tropp2011freedman}]\label{prop:matrix-freedman}
Consider a matrix martingale $\{ \:Y_k : k = 0, 1, 2, \dots \}$ whose values are self-adjoint matrices with dimension $d$, and let $\{ \:X_k : k = 1, 2, 3, \dots \}$ be the difference sequence.  Assume that the difference sequence is uniformly bounded in the sense that
\begin{align*}
 \normsmall{\:X_k}_2  \leq R
\quad\text{almost surely}
\quad\text{for $k = 1, 2, 3, \dots$}.
\end{align*}
Define the predictable quadratic variation process of the martingale as
\begin{align*}
\:{W}_k \eqdef \sum_{j=1}^k \expectedvalue \left[ \:X_j^2 \condbar \{\:X_{s}\}_{s=0}^{j-1} \right],
\quad\text{for $k = 1, 2, 3, \dots$}.
\end{align*}
Then, for all $\varepsilon \geq 0$ and $\sigma^2 > 0$,
\begin{align*}
\probability\left( \exists k \geq 0 : \normsmall{\:Y_k}_2 \geq \varepsilon \ \cap\ 
        \normsmall{ \:W_{k} } \leq \sigma^2 \right)
	\leq 2d \cdot \exp \left\{ - \frac{ \varepsilon^2/2 }{\sigma^2 + R\varepsilon/3} \right\}.
\end{align*}
\end{proposition}

In order to use the previous inequality, we develop the probability of error for any fixed $t$ as
\begin{align*}
&\probability\left( \normsmall{\wh{\:Y}_{\{t,m,N\}}} \geq \varepsilon\right)\\
&\leq \probability\left( \normsmall{\:Y_{\{t,m,N\}}} \geq \varepsilon\right)\\
&= \probability\left( \normsmall{\:Y_{\{t,m,N\}}} \geq \varepsilon \cap \normsmall{\:W_{\{t,m,N\}}} \leq \sigma^2\right)
+ \probability\left( \normsmall{\:Y_{\{t,m,N\}}} \geq \varepsilon \cap \normsmall{\:W_{\{t,m,N\}}} \geq \sigma^2\right)\\
&\leq \underbrace{\probability\left( \normsmall{\:Y_{\{t,m,N\}}} \geq \varepsilon \cap \normsmall{\:W_{\{t,m,N\}}} \leq \sigma^2\right)}_{\mbox{(a)}}
    + \underbrace{\probability\left( \normsmall{\:W_{\{t,m,N\}}} \geq \sigma^2\right)}_{\mbox{(b)}}.
\end{align*}
Using the bound on  $\normsmall{\:X_k}_2 $ from  of Step~2, we can directly apply Proposition~\ref{prop:matrix-freedman} to bound $\mbox{(a)}$ for any fixed $\sigma^2$.
To bound the part $\mbox{(b)}$, we see that for any $r\in[1,\ldots,\{m,m,N\}]$ the predictable quadratic variation process takes the form
\begin{align*}
\:W_r =\sum_{\{s,e,j\}\leq r} \expectedvalue \left[ \frac{1}{N^{2}}\left(\frac{z_{s-1,e,j}}{\wb{p}_{s-1,e}} - \frac{z_{s,e,j}}{\wb{p}_{s,e}}\right)^{2}\:v_{e}\:v_{e}^\transp\:v_{e}\:v_{e}^\transp  \condbar \F_{\{s,e,j\}-1} \right].
\end{align*}
To show that its norm is large only with a small probability, we use the following lemma, proved later in Section~\ref{sec:lemma3}.

\begin{lemma}[Low probability of the large norm of the predictable quadratic variation process]\label{lem:prob-dominance}
    \begin{align*}
        \probability\left( \normsmall{\:W_{\{t,m,N\}}} \geq \frac{9\alpha^2 n\log(\kappa n)}{N}\right)
        \leq n \cdot \exp \left\{ - \frac{N}{\alpha^2(n-1)} \right\}
    \end{align*}
\end{lemma}

Combining Proposition \ref{prop:matrix-freedman} with $\sigma^2 = \frac{9\alpha^2 n\log(\kappa n)}{N}$  and Lemma \ref{lem:prob-dominance} we obtain
\begin{align*}
\probability&\left( \normsmall{\wh{\:Y}_{\{t,m,N\}}} \geq \varepsilon\right)\\
&\leq \probability\left( \normsmall{\:Y_{\{t,m,N\}}} \geq \varepsilon \cap \normsmall{\:W_{\{t,m,N\}}} \leq \sigma^2\right)
+ \probability\left( \normsmall{\:W_{\{t,m,N\}}} \geq \sigma^2\right)\\
&\leq 2n \cdot \exp \left\{ - \frac{ \varepsilon^2/2 }{ \frac{9\alpha^2 n\log(\kappa n)}{N}+ \frac{\varepsilon\alpha^2 (n-1)}{3N}} \right\}
+ n \cdot \exp \left\{ - \frac{N}{\alpha^2(n-1)} \right\}\\
&\leq 2n \cdot \exp \left\{ - \frac{ \varepsilon^2N}{2(9 + \varepsilon/3)\alpha^2 n\log(\kappa n)}\right\}
+ n \cdot \exp \left\{ - \frac{N}{\alpha^2(n-1)}\right\}\\
&\leq 3n \cdot \exp \left\{ - \frac{ \varepsilon^2N }{ 20\alpha^2 n\log(\kappa n)} \right\}\cdot
\end{align*}

Finally, choosing $N$ as in Algorithm \ref{alg:kl_sequence} we obtain the desired result
\begin{align*}
\probability&\left( \normsmall{\wh{\:Y}_{\{t,m,N\}}} \geq \varepsilon\right)
\leq 3n \cdot \exp \left\{ - \frac{ \varepsilon^2N }{ 20\alpha^2 n\log(\kappa n)} \right\}
= 3n \cdot \exp \left\{ - \frac{2\log^2(3\kappa m/\delta)}{\log(\kappa n)} \right\}\\
&\leq 3n \cdot \exp \left\{ - 2\log(3\kappa m/\delta) \right\}
= 3n \left(\frac{\delta}{3\kappa m}\right)^{2}
\leq \frac{\delta}{3\kappa m} \leq\frac{\delta}{2m} 
\end{align*}

\makeatletter{}
\vspace{-0.05in}
\section{Proof of Lemma \ref{lem:prob-dominance} (bound on predictable quadratic variation)}\label{sec:lemma3}
\vspace{-0.05in}

\newcounter{cnt-lem-quad-variation}
\setcounter{cnt-lem-quad-variation}{1}

\textbf{Step \arabic{cnt-lem-quad-variation}\stepcounter{cnt-lem-quad-variation} (a preliminary bound).}
We start by rewriting $\:W_{r}$ as
\begin{align*}
\:W_r =\frac{1}{N^{2}}\sum_{\{s,e,j\}\leq r} \expectedvalue \left[\left(\frac{z_{s-1,e,j}}{\wb{p}_{s-1,e}} - \frac{z_{s,e,j}}{\wb{p}_{s,e}}\right)^{2} \condbar \F_{\{s,e,j\}-1} \right]\:v_{e}\:v_{e}^\transp\:v_{e}\:v_{e}^\transp.
\end{align*}
We rewrite the expectation terms in the equation above as
\begin{align*}
\expectedvalue &\left[ 
\left(\frac{z_{s-1,e,j}}{\wb{p}_{s-1,e}} - \frac{z_{s,e,j}}{\wb{p}_{s,e}}\right)^{2} \condbar \F_{\{s,e,j\}-1} \right]\\
&= \expectedvalue \left[\frac{z_{s-1,e,j}^2}{\wb{p}_{s-1,e}^2} -2 \frac{z_{s-1,e,j}}{\wb{p}_{s-1,e}}\frac{z_{s,e,j}}{\wb{p}_{s,e}} +\frac{z_{s,e,j}^2}{\wb{p}_{s,e}^2} \condbar \F_{\{s,e,j\}-1} \right]\\
&\stackrel{(a)}{=} \expectedvalue \left[\frac{z_{s-1,e,j}^2}{\wb{p}_{s-1,e}^2} -2 \frac{z_{s-1,e,j}}{\wb{p}_{s-1,e}}\frac{z_{s,e,j}}{\wb{p}_{s,e}} +\frac{z_{s,e,j}^2}{\wb{p}_{s,e}^2} \condbar \F_{\{s-1,m,N\}} \right]\\
&= \frac{z_{s-1,e,j}^2}{\wb{p}_{s-1,e}^2} -2 \frac{z_{s-1,e,j}}{\wb{p}_{s-1,e}}\frac{1}{\wb{p}_{s,e}}\expectedvalue \left[z_{s,e,j}\condbar \F_{\{s-1,m,N\}} \right] +\frac{1}{\wb{p}_{s,e}^2}\expectedvalue \left[z_{s,e,j}^2 \condbar \F_{\{s-1,m,N\}} \right]\\
&\stackrel{(b)}{=} \frac{z_{s-1,e,j}}{\wb{p}_{s-1,e}^2}
-2 \frac{z_{s-1,e,j}}{\wb{p}_{s-1,e}}\frac{z_{s-1,e,j}}{\wb{p}_{s-1,e}}
+\frac{1}{\wb{p}_{s,e}^2}\expectedvalue \left[z_{s,e,j}\condbar \F_{\{s-1,m,N\}} \right]\\
&=\frac{1}{\wb{p}_{s,e}^2}\expectedvalue \left[z_{s,e,j} \condbar \F_{\{s-1,m,N\}} \right] - \frac{z_{s-1,e,j}}{\wb{p}_{s-1,e}^2}\\
&\stackrel{(c)}{=}\frac{1}{\wb{p}_{s,e}}\frac{z_{s-1,e,j}}{\wb{p}_{s-1,e}} - \frac{z_{s-1,e,j}}{\wb{p}_{s-1,e}^2}
=\frac{z_{s-1,e,j}}{\wb{p}_{s-1,e}}\left(\frac{1}{\wb{p}_{s,e}} - \frac{1}{\wb{p}_{s-1,e}}\right),
\end{align*}
where in $(a)$ we use the fact that the approximate probabilities $\wb{p}_{s-1,e}$ and $\wb{p}_{s,e}$ and $z_{s-1,e,j}$ are fixed at the end of the previous iteration, while in $(b)$ and $(c)$ we use the fact that $z_{s,e,j}$ is a Bernoulli of parameter $\wb{p}_{s,e}/\wb{p}_{s-1,e}$ (whenever $z_{s-1,e,j}$ is equal to 1).
Therefore we can write $\:W$ at the end of any iteration $t$ as
\begin{align*}
\:W_{\{t,m,N\}} &= \frac{1}{N^{2}}\sum_{j=1}^{N} \sum_{e=1}^m \sum_{s=1}^t \frac{z_{s-1,e,j}}{\wb{p}_{s-1,e}}\left(\frac{1}{\wb{p}_{s,e}} - \frac{1}{\wb{p}_{s-1,e}}\right)\:v_{e}\:v_{e}^\transp\:v_{e}\:v_{e}^\transp.
\end{align*}

We can now upper-bound $\:W$ as
\begin{align*}
\:W_{\{t,m,N\}} &= \frac{1}{N^2}\sum_{j=1}^{N}\sum_{e=1}^m\:v_{e}\:v_{e}^\transp\:v_{e}\:v_{e}^\transp\sum_{s=1}^t\frac{z_{s-1,e,j}}{\wb{p}_{s-1,e}}\left(\frac{1}{\wb{p}_{s,e}} - \frac{1}{\wb{p}_{s-1,e}}\right)\\
&\preceq \frac{1}{N^2}\sum_{j=1}^{N}\sum_{e=1}^m\:v_{e}\:v_{e}^\transp\:v_{e}\:v_{e}^\transp\left(\max_{s=1\dots t}\left\{\frac{z_{s-1,e,j}}{\wb{p}_{s-1,e}}\right\}\right)\sum_{s=1}^t\left(\frac{1}{\wb{p}_{s,e}} - \frac{1}{\wb{p}_{s-1,e}}\right)\\
&= \frac{1}{N^2} \sum_{j=1}^{N}\sum_{e=1}^m\:v_{e}\:v_{e}^\transp\:v_{e}\:v_{e}^\transp\left(\max_{s=1\dots t}\left\{\frac{z_{s-1,e,j}}{\wb{p}_{s-1,e}}\right\}\right)\left(\frac{1}{\wb{p}_{t,e}} - 1\right)\\
&\preceq \frac{1}{N^2}\sum_{j=1}^{N}\sum_{e=1}^m\frac{1}{\wb{p}_{t,e}}\:v_{e}\:v_{e}^\transp\:v_{e}\:v_{e}^\transp\left(\max_{s=1\dots t}\left\{\frac{z_{s-1,e,j}}{\wb{p}_{s-1,e}}\right\}\right),
\end{align*}
where in the first inequality we use the fact that $\left(\frac{1}{\wb{p}_{s,e}} - \frac{1}{\wb{p}_{s-1,e}}\right)$
and $\frac{z_{s-1,e,j}}{\wb{p}_{s-1,e}}$ are always positive and the
definition of L{\"o}wner's ordering and in the second inequality we just upper-bound the last term by $1/\wb{p}_{t,e}$. We  further bound the previous term by studying $\wb{p}_{t,e}$. If $\normsmall{\:Y_{\{t-1,m,N\}}} \leq \varepsilon$, then $\wb{p}_{t,e} = \wt{p}_{t,e}$ and by Proposition~\ref{prop:alpha.good}, $\wb{p}_{t,e}$ is $\alpha$-accurate and hence 
\begin{align*}
    \frac{1}{\wb{p}_{t,e}} = \frac{1}{\wt{p}_{t,e}} \leq \frac{\alpha^2}{p_{t,e}} \leq \frac{\alpha^2}{p_{m,e}}\CommaBin
\end{align*}
where we use the fact that probabilities (as much as effective resistances) are non-increasing over iterations. Furthermore, in the case if $\normsmall{\:Y_{\{t-1,m,N\}}} > \varepsilon$, let $s<t$ be the last time when $\normsmall{\:Y_{\{s,m,N\}}} \leq \varepsilon$. In this case, $\wb{p}_{t,e} = \wb{p}_{s,e}$, since the process is ``frozen'' and therefore,
\begin{align*}
    \frac{1}{\wb{p}_{t,e}} = \frac{1}{\wb{p}_{s,e}} \leq \frac{\alpha^2}{p_{s,e}} \leq \frac{\alpha^2}{p_{m,e}}\CommaBin
\end{align*}
since at iteration $s$, the probability $\wb{p}_{s,e}$ is guaranteed to be $\alpha$-accurate.
As a result, we obtain
\begin{align*}
\:W_{\{t,m,N\}} 
&\preceq \frac{1}{N^2}\sum_{j=1}^{N}\sum_{e=1}^m\frac{1}{\wb{p}_{t,e}}\:v_{e}\:v_{e}^\transp\:v_{e}\:v_{e}^\transp\left(\max_{s=1\dots t}\left\{\frac{z_{s-1,e,j}}{\wb{p}_{s-1,e}}\right\}\right)\\
& \preceq \frac{1}{N^2}\sum_{j=1}^{N}\sum_{e=1}^m\frac{\alpha^2}{p_{m,e}}\:v_{e}\:v_{e}^\transp\:v_{e}\:v_{e}^\transp\left(\max_{s=1\dots t}\left\{\frac{z_{s-1,e,j}}{\wb{p}_{s-1,e}}\right\}\right)\cdot
\end{align*}
Using the definition of $p_{m,e} = a_e r_{m,e}/(n-1)$ and $\:v_e = \sqrt{a_e}\:L_{\Gg}^{-1/2} \:b_e$ we have $\:v_e^\transp \:v_e = a_e r_{m,e}$ and thus
\begin{align}
\:W_{\{t,m,N\}}
& \preceq \frac{\alpha^2}{N^2}\sum_{j=1}^{N}\sum_{e=1}^m\frac{1}{p_{m,e}}\:v_{e}\:v_{e}^\transp\:v_{e}\:v_{e}^\transp\left(\max_{s=1\dots t}\left\{\frac{z_{s-1,e,j}}{\wb{p}_{s-1,e}}\right\}\right)\nonumber\\
&= \frac{\alpha^2}{N^2}\sum_{j=1}^{N}\sum_{e=1}^m\frac{(n-1)a_{e} r_{m,e}}{a_e r_{m,e}}\:v_{e}\:v_{e}^\transp\left(\max_{s=1\dots t}\left\{\frac{z_{s-1,e,j}}{\wb{p}_{s-1,e}}\right\}\right)\nonumber\\
&= \frac{\alpha^2(n-1)}{N^2}\sum_{j=1}^{N}\sum_{e=1}^m\:v_{e}\:v_{e}^\transp\left(\max_{s=1\dots t}\left\{\frac{z_{s-1,e,j}}{\wb{p}_{s-1,e}}\right\}\right)\cdot\label{eq:dominance-W}
\end{align}

\textbf{Step \arabic{cnt-lem-quad-variation}\stepcounter{cnt-lem-quad-variation} (introduction of a stochastically dominant process).}
We want to study
\begin{align*}
\max_{s=1\dots t}\frac{z_{s-1,e,j}}{\wb{p}_{s-1,e}}  = \max_{s=0\dots t-1}\frac{z_{s,e,j}}{\wb{p}_{s,e}}\cdot
\end{align*}
We know trivially that this quantity is larger or equal than 1 because $z_{0,e,j}/\wb{p}_{0,e} = 1$,
but upper-bounding this quantity is not trivial as the evolution
of the various $\wb{p}_{s,e}$ depends in a complex way on the interaction
between the random variables $z_{s,e,j}$.
To simplify the following analysis, we introduce the random variable $z_{t+1,e,j}/\wb{p}_{t+1,e}$,
where $\wb{p}_{t+1,e}$ is deterministically defined as $\wb{p}_{t+1,e} = p_{t,e}/\alpha^2$
and $z_{t+1,e,j}$ is defined, coherently with the other $z_{s,e,j}$, as 0 if $z_{t,e,j} = 0$ or as the result of a Bernoulli
flip with probability $\wb{p}_{t+1,e}/\wb{p}_{t,e}$ if $z_{t,e,j} = 1$.
We have that
\begin{align*}
 \max_{s=0\dots t-1}\frac{z_{s,e,j}}{\wb{p}_{s,e}} \leq  \max_{s=0\dots t}\frac{z_{s,e,j}}{\wb{p}_{s,e}} \leq  \max_{s=0\dots t+1}\frac{z_{s,e,j}}{\wb{p}_{s,e}}\CommaBin
\end{align*}
but now we know exactly the value of the last $\wb{p}_{t+1,e}$.
Morevoer, we did not lose much since the bounds are tight for
the (realizable) worst case when $\wb{p}_{t-1,e} =\wb{p}_{t,e} =  \wb{p}_{t+1,e} = p_{t,e}$.
Finding a deterministic bound on $\max_{s=0\dots t+1}\frac{z_{s,e,j}}{\wb{p}_{s,e}}$ is not easy,
since $\wb{p}_{t+1,e} = p_{t,e}$ can be very close to
0.
Nonetheless, whenever $\wb{p}_{s,e}$ is significantly
smaller than $\wb{p}_{s-1,e}$, the probability of keeping a copy of edge $e$ at
iteration $s$ (i.e., $z_{s,e,j}=1$) is also very small. As a result, we expect
the ratio $\frac{z_{s,e,j}}{\wb{p}_{s,e}}$ to be still small with
high probability. Unfortunately, due to the dependency between different copies
of the edges at different iterations, it seems difficult to exploit this intuition directly
to provide an overall high-probability bound on $\:W_{\{t,m,N\}}$. For this
reason, we simplify the analysis by replacing each of the (potentially
dependent) chains $\{z_{s,e,j}/\wb{p}_{s,e}\}_{s=0}^{t+1}$ with a set of
(independent) random variables $w_{0,e,j}$ that will stochastically dominate
them.

We define the random variable $w_{s,e,j}$ using the following conditional
distribution\footnote{
Notice that unlike $z_{s,e,j}$, $w_{s,e,j}$ is no longer $\F_{\{s,m,N\}}$-measurable but it is $\F'_{\{s,m,N\}}$-measurable, where 
\begin{align*}
\F'_{\{s,e,j\}} = \left\{ u_{s',e',j'},\; \forall\{s',e',j'\} \leq \{s,e,j\} \right\} \cup \left\{  w_{s,e,j} \right\}
= \F_{\{s,e,j\}} \cup \left\{  w_{s,e,j} \right\}.
\end{align*}
} 
\begin{align*}
\probability\left(\frac{1}{w_{s,e,j}} \leq a \condbar \F_{\{s,m,N\}}\right)
 = \begin{cases}
0 &\text{ for }\quad a < 1/\wb{p}_{s,e}\\
1-\frac{1}{\wb{p}_{s,e}a} &\text{ for }\quad 1/\wb{p}_{s,e} \leq a < \alpha^2/p_{t,e}\\
1 &\text{ for }\quad \alpha^2/p_{t,e} \leq a
\end{cases}.
\end{align*}
Note that the distribution of $\frac{1}{w_{s,e,j}}$ conditioned on $\F_{\{s,m,N\}}$
is determined by only $\wb{p}_{s,e}$, $p_{t,e}$, and~$\alpha$, where~$p_{t,e}$ and~$\alpha$
are fixed. Remembering that $\wb{p}_{s,e}$ is a function of
$\F_{\{s-1,m,N\}}$ (computed using the previous iteration),
we have that  
\begin{align*}
\probability\left(\frac{1}{w_{s,e,j}} \leq a \condbar \F_{\{s,m,N\}}\right)
= \probability\left(\frac{1}{w_{s,e,j}} \leq a \condbar \F_{\{s-1,m,N\}}\right).
    \end{align*}
Notice that  in the definition of $w_{s,e,j}$, none of the other $w_{s',e',j'}$
(for any different $s'$, $e'$, or $j'$) appears
and $\wb{p}_{s,e}$ is a function of
$\F_{\{s-1,m,N\}}$. It follows that given  $\F_{\{s-1,m,N\}}$, $w_{s,e,j}$ is independent from all other $w_{s',e',j'}$
(for any different $s'$, $e'$, or $j'$).
\begin{figure}[t]\label{fig:rand-var-dep-graph}
\includegraphics[width=0.8\textwidth]{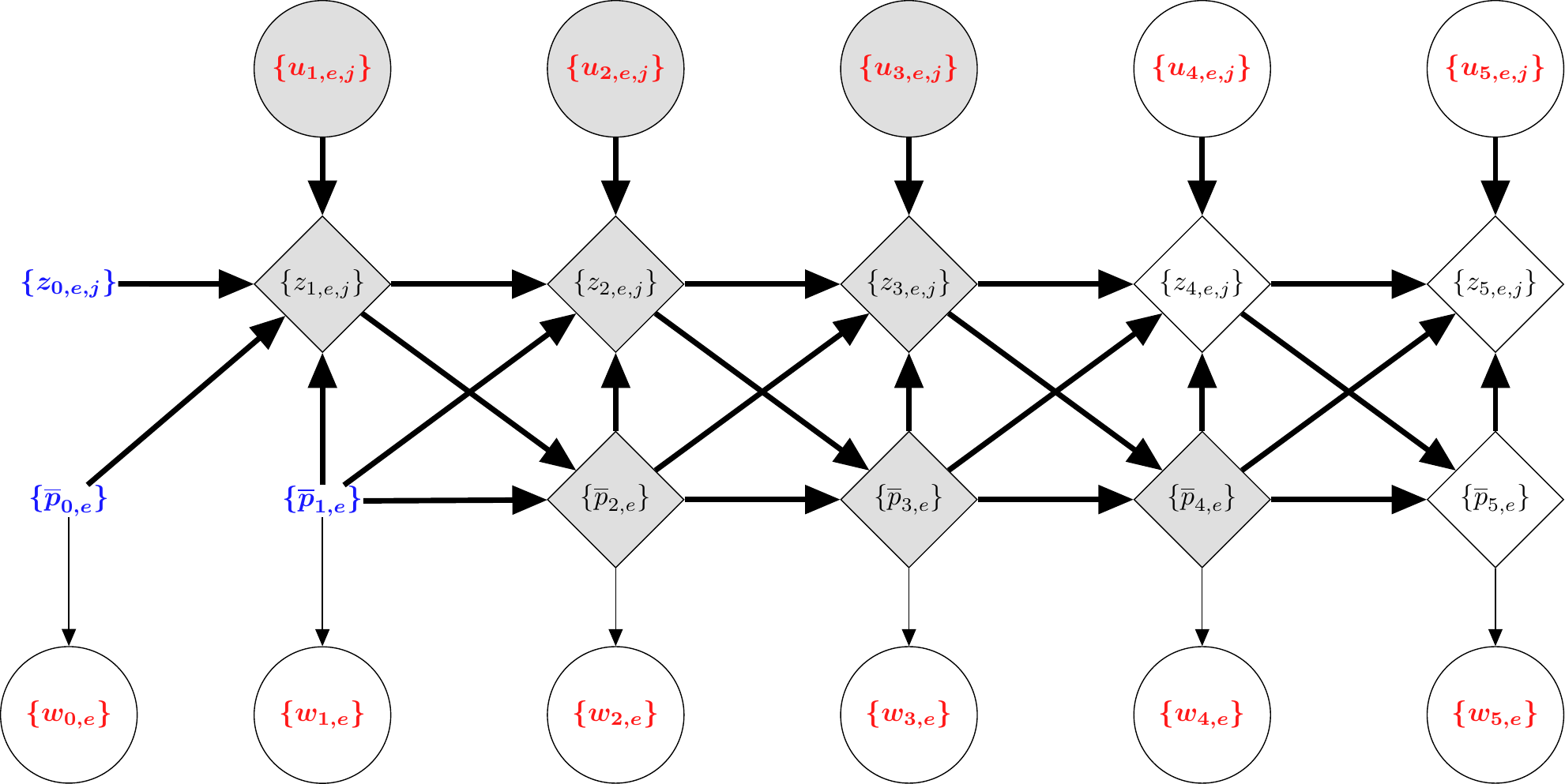}
\caption{The dependence graph of the considered variables. \rcolb{Red} variables are \rcolb{random}, Black variables are deterministically computed using their input (a function of their input), with bold lines indicating the deterministic (functional) relation. \bcolb{Blue} variables are \bcolb{constants}. A \colorbox{gray!30}{grey filling} indicates that a random variable is \colorbox{gray!30}{observed}, or a function of observed variables.}
\label{fig:rand-var-dep-graph}
\end{figure}
This is easier to see in the probabilistic graphical model reported in
Figure \ref{fig:rand-var-dep-graph}, which illustrates the dependence between
the various variables.

Finally for  the special case $w_{0,e,j}$ the definition above reduces to
\begin{align}\label{eq:distro.w}
\probability\left(\frac{1}{w_{0,e,j}} \leq a\right)
 = \begin{cases}
0 &\text{ for }\quad a < 1\\
1-\frac{1}{a} &\text{ for }\quad 1 \leq a < \alpha^2/p_{t,e}\\
1 &\text{ for }\quad \alpha^2/p_{t,e} \leq a
\end{cases},
\end{align}
since $\wb{p}_{0,e}=1$ by definition.
From this definition, $w_{0,e,j}$ and $w_{0,e',j'}$ are all
independent, and this will allow us to use stronger concentration inequalities
for independent random variables.

We remind the reader that a random variable $A$ stochastically dominates random
variable $B$, if for all values $a$ the two equivalent conditions are verified
\begin{align*}
\probability(A \geq a) \geq \probability(B \geq a) \Leftrightarrow \probability(A \leq a) \leq \probability(B \leq a).
\end{align*}
As a consequence, if $A$ dominates $B$ the following implication holds
\begin{align*}
\probability(A \geq a) \geq \probability(B \geq a) \implies \expectedvalue[A] \geq \expectedvalue[B],
\end{align*}
while the reverse ($A$ dominates $B$ if $\expectedvalue[A] \geq \expectedvalue[B]$) is not
true in general.
Following this definition of  stochastic dominance, our goal is to prove
\begin{align*}
\probability\left(\max_{s=0\dots t+1}\frac{z_{s,e,j}}{\wb{p}_{s,e}} \leq a\right)
\geq 
\probability\left(\frac{1}{w_{0,e,j}} \leq a \right).
\end{align*}
We prove this inequality by proceeding backward with a sequence of conditional probabilities.

\textbf{Step \arabic{cnt-lem-quad-variation}\stepcounter{cnt-lem-quad-variation} (case $t+1$).}
We first study the distribution of the maximum conditional to the state of the algorithm at the end of iteration $t$, i.e., $\Fp_{\{t,m,N\}}$, that is
\begin{align*}
\rho_a \eqdef \probability\left(\max\left\{ \max_{s=0 \dots t}\frac{z_{s,e,j}}{\wb{p}_{s,e}} ; \frac{z_{t+1,e,j}}{\wb{p}_{t+1,e}}\right\} \leq a \condbar \Fp_{\{t,m,N\}}\right).
\end{align*}
We distinguish two cases, depending on whether $z_{t,e,j} = 0$ or $1$, which directly depends on $\Fp_{\{t,m,N\}}$.
\begin{itemize}
\item If $z_{t,e,j} = 0$, then by definition $z_{t+1,e,j} = 0$ and thus for any $a$
\begin{align*}
\rho_a = \probability\left( \max_{s=0 \dots t-1}\frac{z_{s,e,j}}{\wb{p}_{s,e}} \leq a \condbar \Fp_{\{t,m,N\}}\right),
\end{align*}
since the maximum is always attained by one of the terms for which $z_{s,e,j} = 1$.
\item If $z_{t,e,j} = 1$, then by definition $z_{t+1,e,j} \big| \Fp_{\{t,m,N\}}$ is a Bernoulli variable of parameter $\wb{p}_{e,t+1}/\wb{p}_{e,t}$. Since $z_{t,e,j}=1$ and $\wb{p}_{t,e} \leq \wb{p}_{s,e}$ for all $s\leq t$, the inner maximum in $\rho_a$ is always attained by $\frac{z_{t,e,j}}{\wb{p}_{t,e}}$ and thus we can write $\rho_a$ as
\begin{align*}
\rho_a = \probability\left( \max\left\{\frac{1}{\wb{p}_{t,e}}; \frac{z_{t+1,e,j}}{\wb{p}_{t+1,e}}\right\} \leq a \condbar \Fp_{\{t,m,N\}}\right).
\end{align*}
We can easily study the c.d.f. of the maximum depending on the value of $a$ as follows:
\begin{itemize}
\item If $a < 1/\wb{p}_{t,e}$, then $\rho_a = 0$, since the maximum is at least $\frac{1}{\wb{p}_{t,e}}$.
\item If $1/\wb{p}_{t,e} \leq a < 1/\wb{p}_{t+1,e}$, we are in the case where the Bernoulli variable takes value $0$, and thus
\begin{align*}
\rho_a = \probability\left( z_{t+1,e,j}=0 \condbar \Fp_{\{t,m,N\}}\right) = 1 - \frac{\wb{p}_{t+1,e}}{\wb{p}_{t,e}}\cdot
\end{align*}
\item If $a \geq 1/\wb{p}_{t+1,e}$, then $\rho_a=1$, since $a$ is bigger or equal than the largest value attainable by the maximum.
\end{itemize}
\end{itemize}

We now move to analyze the distribution
\begin{align*}
\pi_a \eqdef \probability\left(\max\left\{ \max_{s=0 \dots t}\frac{z_{s,e,j}}{\wb{p}_{s,e}} ; \frac{z_{t,e,j}}{w_{t,e,j}}\right\} \leq a \condbar \Fp_{\{t,m,N\}}\right),
\end{align*}
where we replaced $z_{t+1,e,j}/\wb{p}_{t+1,e}$  with the variable $z_{t,e,j}/w_{t,e,j}$. Note that in this expression only $w_{t,e,j}$ is random, and all other terms
are fixed given $\Fp_{\{t,m,N\}}$. If $z_{t,e,j} = 0$, then for any $a$,
\begin{align*}
\pi_a = \probability\left( \max_{s=0 \dots t-1}\frac{z_{s,e,j}}{\wb{p}_{s,e}} \leq a \condbar \Fp_{\{t,m,N\}}\right).
\end{align*}

\begin{figure}[t]
\includegraphics[width=0.6\textwidth]{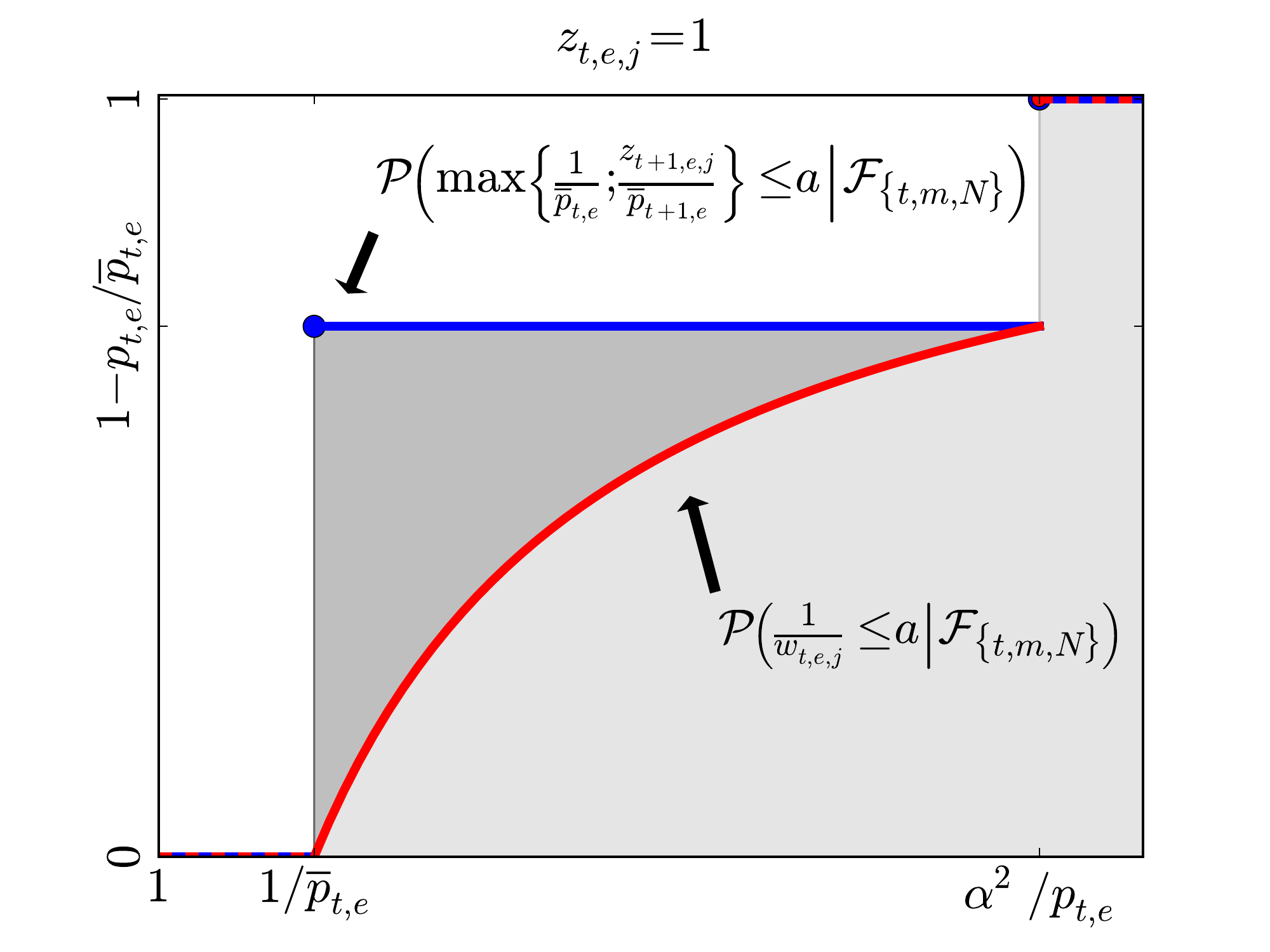}
\caption{C.d.f.\,of $\max\left\{ 1/\wb{p}_{t,e} ; z_{t+1,e,j}/\wb{p}_{t+1,e}\right\}$ and
$1/w_{t,e,j}$ conditioned on $\mathcal{F}_{\{t,m,N\}}$}\label{fig:dominance}
\end{figure}

On the other hand, if $z_{t,e,j} = 1$, we can use the definition of $w_{t,e,j}$ to determine its c.d.f.. As a result, we notice that if $z_{t,e,j}=0$, then $\rho_a=\pi_a$ for any $a$, while for $z_{t,e,j}=1$ we have

\begin{align*}
\probability&\left(\max\left\{ \max_{s=0 \dots t}\frac{z_{s,e,j}}{\wb{p}_{t,e}} ; \frac{z_{t+1,e,j}}{\wb{p}_{t+1,e}}\right\} \leq a \condbar \Fp_{\{t,m,N\}}\right)\\
&=
\begin{cases}
0 &\text{ for }\quad 1/\wb{p}_{t,e} < a \text{ and } z_{t,e,j} = 1\\
1-\frac{\wb{p}_{t+1,e}}{\wb{p}_{t,e}} &\text{ for }\quad 1/\wb{p}_{t,e} \leq a < 1/\wb{p}_{t+1,e} \text{ and } z_{t,e,j} = 1\\
1 &\text{ for }\quad 1/\wb{p}_{t+1,e} \leq a \text{ and } z_{t,e,j} = 1
\end{cases}\\
&\geq
\begin{cases}
0 &\text{ for }\quad 1/\wb{p}_{t,e} < a \text{ and } z_{t,e,j} = 1\\
1-\frac{1}{\wb{p}_{t,e}a} &\text{ for }\quad 1/\wb{p}_{t,e} \leq a < 1/\wb{p}_{t+1,e} \text{ and } z_{t,e,j} = 1\\
1 &\text{ for }\quad 1/\wb{p}_{t+1,e} = \alpha^2/p_{t,e} \leq a  \text{ and } z_{t,e,j} = 1
\end{cases}\\
&=
\probability\left(\max\left\{ \max_{s=0 \dots t}\frac{z_{s,e,j}}{\wb{p}_{t,e}} ; \frac{z_{t,e,j}}{w_{t,e,j}}\right\} \leq a \condbar \Fp_{\{t,m,N\}}\right),
\end{align*}where we used the definition $\wb{p}_{t+1,e} = p_{t,e}/\alpha^2$,
and the fact that since $\wb{p}_{t,e}$ is always either $\alpha$-accurate,
or frozen to a previous $\alpha$-accurate value $\wt{p}_{s,e}$, we have that
$\wb{p}_{t,e} \geq \wb{p}_{t+1,e} = p_{t,e}/\alpha^2$. The inequality between
$\rho_a$ and $\pi_a$ comes from the fact that in the intermediate interval $a$
is such that $a < 1/\wb{p}_{t+1,e}$ and then $1-\frac{1}{\wb{p}_{t,e}a} \leq
1-\frac{\wb{p}_{t+1,e}}{\wb{p}_{t,e}}$, as it is easy to observe in
Fig.~\ref{fig:dominance}. We can thus conclude that $\rho_a \geq
\pi_a$ for any $a$.

Now that we introduced $w_{t,e,j}$ in the maximum of the conditional probability, we can focus on the overall probability as
\begin{align*}
&\probability\left(\max_{s=0 \dots t+1}\frac{z_{s,e,j}}{\wb{p}_{s,e}} \leq a\right)
=\sum_{\Fp_{\{t,m,N\}}}\probability\left(\max_{s=0 \dots t+1}\frac{z_{s,e,j}}{\wb{p}_{s,e}} \leq a \condbar \Fp_{\{t,m,N\}}\right)\probability(\Fp_{\{t,m,N\}})\\
&=\sum_{\Fp_{\{t,m,N\}}}\probability\left(\max\left\{ \max_{s=0 \dots t}\frac{z_{s,e,j}}{\wb{p}_{s,e}} ; \frac{z_{t+1,e,j}}{\wb{p}_{t+1,e}}\right\} \leq a \condbar \Fp_{\{t,m,N\}}\right)\probability(\Fp_{\{t,m,N\}})\\
&\geq \sum_{\Fp_{\{t,m,N\}}} \probability\left(\max\left\{ \max_{s=0 \dots t}\frac{z_{s,e,j}}{\wb{p}_{s,e}} ; \frac{z_{t,e,j}}{w_{t,e,j}}\right\} \leq a \condbar \Fp_{\{t,m,N\}}\right)\probability(\Fp_{\{t,m,N\}})\\
&= \probability\left(\max\left\{ \max_{s=0 \dots t}\frac{z_{s,e,j}}{\wb{p}_{s,e}} ; \frac{z_{t,e,j}}{w_{t,e,j}}\right\} \leq a \right).
\end{align*}\textbf{Step \arabic{cnt-lem-quad-variation}\stepcounter{cnt-lem-quad-variation} (generic case).}
We now proceed by peeling off layers from the end of the chain one by one. We show how to move from an iteration $k\leq t$ to $k-1$.
{\footnotesize
\begin{align}\label{eq:generic.case}
& \probability\left(\max\left\{ \max_{s=0 \dots k}\frac{z_{s,e,j}}{\wb{p}_{s,e}} ; \frac{z_{k,e,j}}{w_{k,e,j}}\right\} \leq a \right)
=\probability\left(\max\left\{ \max_{s=0 \dots k-1}\frac{z_{s,e,j}}{\wb{p}_{s,e}} ;\frac{z_{k,e,j}}{\wb{p}_{k,e}}; \frac{z_{k,e,j}}{w_{k,e,j}}\right\} \leq a \right)\\
&=\sum_{\Fp_{\{k-1,m,N\}}}\probability\left(\max\left\{ \max_{s=0 \dots k-1}\frac{z_{s,e,j}}{\wb{p}_{s,e}} ;\frac{z_{k,e,j}}{\wb{p}_{k,e}}; \frac{z_{k,e,j}}{w_{k,e,j}}\right\} \leq a \condbar \Fp_{\{k-1,m,N\}} \right)\probability(\Fp_{\{k-1,m,N\}})\nonumber\\
&=\sum_{\Fp_{\{k-1,m,N\}}}\probability\left(\max\left\{ \max_{s=0 \dots k-1}\frac{z_{s,e,j}}{\wb{p}_{s,e}} ;z_{k,e,j}\max\left\{\frac{1}{\wb{p}_{k,e}}; \frac{1}{w_{k,e,j}}\right\}\right\} \leq a \condbar \Fp_{\{k-1,m,N\}} \right)\probability(\Fp_{\{k-1,m,N\}})\nonumber\\
&=\sum_{\Fp_{\{k-1,m,N\}}}\probability\left(\max\left\{ \max_{s=0 \dots k-1}\frac{z_{s,e,j}}{\wb{p}_{s,e}} ;z_{k-1,e,j}\indfunc\left\{u_{k,e,j} \leq \frac{\wb{p}_{k,e}}{\wb{p}_{k-1,e}}\right\}\frac{1}{w_{k,e,j}}\right\} \leq a \condbar \Fp_{\{k-1,m,N\}} \right)\probability(\Fp_{\{k-1,m,N\}}),\nonumber
\end{align}}where in the last equality we use the fact that the inner maximum is always attained by $1/w_{k,e,j}$ since by definition $1/w_{k,e,j}$ is lower-bounded by $1/\wb{p}_{k,e}$ and we use the recursive definition of $z_{k,e,j}$. Since~$u_{k,e,j}$ and
$w_{k,e,j}$ are independent given $\Fp_{\{k-1,m,N\}}$, we have
\begin{align}
\probability&\left( \indfunc\left\{u_{k,e,j} \leq \frac{\wb{p}_{k,e}}{\wb{p}_{k-1,e}}\right\}\frac{1}{w_{k,e,j}} \leq a \condbar \Fp_{\{k-1,m,N\}}\right)\nonumber\\
&=
\begin{cases}
0 &\text{ for }\quad a \leq 0 \nonumber\\
1 - \frac{\wb{p}_{k,e}}{\wb{p}_{k-1,e}} &\text{ for }\quad  0 \leq a < 1/\wb{p}_{k,e}\\
1-\frac{\wb{p}_{k,e}}{\wb{p}_{k-1,e}} +\frac{\wb{p}_{k,e}}{\wb{p}_{k-1,e}}\left(1-\frac{1}{\wb{p}_{k,e}a}\right) = 1-\frac{1}{\wb{p}_{k-1,e}a}  &\text{ for }\quad 1/\wb{p}_{k,e} \leq a < \alpha^2/p_{t,e} \\ 
1 &\text{ for }\quad \alpha^2/p_{t,e} \leq a
\end{cases}\nonumber\\
&\geq 
\begin{cases}
0 &\text{ for }\quad a < 1/\wb{p}_{k-1,e} \\
1-\frac{1}{\wb{p}_{k-1,e}a} &\text{ for }\quad  1/\wb{p}_{k-1,e} \leq a < 1/\wb{p}_{k,e} \\
1-\frac{1}{\wb{p}_{k-1,e}a}  &\text{ for }\quad 1/\wb{p}_{k,e} \leq a < \alpha^2/p_{t,e}  \\
1 &\text{ for }\quad \alpha^2/p_{t,e}  \leq a 
\end{cases}\label{w.stoch123}\\
&=\probability\left(\frac{1}{w_{k-1,e,j}} \leq a \condbar \Fp_{\{k-2,m,N\}}\right)
=\probability\left(\frac{1}{w_{k-1,e,j}} \leq a \condbar \Fp_{\{k-1,m,N\}}\right) \nonumber.
\end{align}
Since given $\F_{\{k-1,m,N\}}$, everything in \eqref{eq:generic.case} is fixed except  $u_{k,e,j}$ and $w_{k,e,j}$, we can use the stochastic dominance in~\eqref{w.stoch123}  to obtain
\begin{align*}
\probability\left(\max\left\{ \max_{s=0 \dots k}\frac{z_{s,e,j}}{\wb{p}_{s,e}} ; \frac{z_{k,e,j}}{w_{k,e,j}}\right\} \leq a \right) \geq \probability\left(\max\left\{ \max_{s=0 \dots k-1}\frac{z_{s,e,j}}{\wb{p}_{s,e}} ; \frac{z_{k-1,e,j}}{w_{k-1,e,j}}\right\} \leq a \right).
\end{align*}
Applying the inequality recursively from $k=t$ to $k=1$ 
removes all $z_{s,e,j}$ from the maximum and we
are finally left with only $w_{0,e,j}$ as we wanted,
\begin{align*}
\probability\left(\max_{s=0 \dots t+1}\frac{z_{s,e,j}}{\wb{p}_{s,e}} \leq a\right)
\geq 
\probability\left(\max\left\{\frac{z_{0,e,j}}{\wb{p}_{0,e}} ; \frac{z_{0,e,j}}{w_{0,e,j}}\right\} \leq a \right)
\geq 
\probability\left(\frac{1}{w_{0,e,j}} \leq a \right),
\end{align*}
where in the last inequality we used that $z_{0,e,j} = 1$ from the definition of the algorithm 
and $\wb{p}_{0,e} = 1$ while $w_{0,e,j} \leq 1$ by~\eqref{eq:distro.w}.

\textbf{Step \arabic{cnt-lem-quad-variation}\stepcounter{cnt-lem-quad-variation} (stochastic dominance on $\:W_{\{t,m,N\}}$).}
Now that we proved the stochastic dominance of $1/w_{0,e,j}$, we plug this result in the definition of $\:W_{\{t,m,N\}}$. For the sake of notation, we introduce the term $\wb{p}_{t+1,e,j}^{\max}$
to indicate the maximum over the first $t+1$ step of copy $e,j$ such that
\begin{align*}
\max_{s=0 \dots t+1}\frac{z_{s,e,j}}{\wb{p}_{s,e}} = \frac{1}{\wb{p}_{t+1,e,j}^{\max}}\cdot
\end{align*}
We first notice that while $\:Y_{\{t,m,N\}}$ is not
necessarily PSD, $\:W_{\{t,m,N\}}$ is a sum of PSD matrices.
Introducing the function $\Lambda(\{1/\wb{p}_{t+1,e,j}^{\max}\}_{e,j})$ we can restate Eq.~\ref{eq:dominance-W} as
\begin{align*}
\normsmall{\:W_{\{t,m,N\}}}
=\lambda_{\max}(\:W_{\{t,m,N\}})
\leq \Lambda(\{1/\wb{p}_{t+1,e,j}^{\max}\}_{e,j}) \eqdef
\lambda_{\max}\left(\frac{\alpha^2(n-1)}{N^2}\sum_{j=1}^{N}\sum_{e=1}^m\frac{1}{\wb{p}_{t+1,e,j}^{\max}}\:v_{e}\:v_{e}^\transp\right).
\end{align*}
In Step 4, we showed that $1/\wb{p}_{t+1,e,j}^{\max}$ is stochastically
dominated by $1/w_{0,e,j}$ for every $e$ and~$j$. In order to bound
$\Lambda(\{1/\wb{p}_{t+1,e,j}^{\max}\}_{e,j})$
we need to show that this dominance also applies to the summation
over all edges inside the matrix norm. Since the matrix $\sum_{j=1}^{N}\sum_{e=1}^m\frac{1}{\wb{p}_{t+1,e,j}^{\max}}\:v_{e}\:v_{e}^\transp$ is symmetric, we
can reformulate $\Lambda(\{1/\wb{p}_{t+1,e,j}^{\max}\}_{e,j})$ as
\begin{align*}
&\lambda_{\max}\left(\frac{\alpha^2(n-1)}{N^2}\sum_{j=1}^{N}\sum_{e=1}^m\frac{1}{\wb{p}_{t+1,e,j}^{\max}}\:v_{e}\:v_{e}^\transp\right)
= \max_{\:x : \normsmall{\:x} = 1} \:x^\transp\left(\frac{\alpha^2(n-1)}{N^2}\sum_{j=1}^{N}\sum_{e=1}^m\frac{1}{\wb{p}_{t+1,e,j}^{\max}}\:v_{e}\:v_{e}^\transp\right)\:x\\
&= \max_{\:x : \normsmall{\:x} = 1}\frac{\alpha^2(n-1)}{N^2}\sum_{j=1}^{N}\sum_{e=1}^m\frac{1}{\wb{p}_{t+1,e,j}^{\max}}\:x^\transp\:v_{e}\:v_{e}^\transp\:x
= \max_{\:x : \normsmall{\:x} = 1}\frac{\alpha^2(n-1)}{N^2}\sum_{j=1}^{N}\sum_{e=1}^m\frac{1}{\wb{p}_{t+1,e,j}^{\max}}\left(\:v_{e}^\transp\:x\right)^2.
\end{align*}
From this reformulation, it is easy to see that the function $\Lambda(\{1/\wb{p}_{t+1,e,j}^{\max}\}_{e,j})$ is
monotonically increasing w.r.t.\,the individual $1/\wb{p}_{t+1,e,j}^{\max}$, or
in other words that increasing an $1/\wb{p}_{t+1,e,j}^{\max}$ without decreasing
the others can only increase the maximum.
Introducing $\Lambda(\{1/w_{0,e,j}\}_{e,j})$ as
\begin{align*}
\Lambda(\{1/w_{0,e,j}\}_{e,j}) \eqdef \max_{\:x : \normsmall{\:x} = 1}\frac{\alpha^2(n-1)}{N^2}\sum_{j=1}^{N}\sum_{e=1}^m\frac{1}{w_{0,e,j}}\left(\:v_{e}^\transp\:x\right)^2.
\end{align*}
we now need to prove the stochastic dominance of $\Lambda(\{1/w_{0,e,j}\}_{e,j})$ over
$\Lambda(\{1/\wb{p}_{t+1,e,j}^{\max}\}_{e,j})$.
Using the definition of $1/\wb{p}_{t+1,e,j}^{\max}$ we have
\begin{align*}
\probability&\left(\Lambda\left(\left\{\frac{1}{\wb{p}_{t+1,e,j}^{\max}}\right\}_{e,j}\right) \leq a\right)
= \probability\left(\Lambda\left(\left\{\max_{s=0  \dots  t+1}\frac{z_{s,e,j}}{\wb{p}_{s,e}}\right\}_{e,j}\right) \leq a\right)\\
& =\probability\left(\Lambda\left(\left\{\max\left\{ \max_{s=0 \dots t}\frac{z_{s,e,j}}{\wb{p}_{s,e}} ; \frac{z_{t+1,e,j}}{\wb{p}_{t+1,e}}\right\}\right\}_{e,j}\right) \leq a \right)\\
&=\sum_{\Fp_{\{t,m,N\}}}\probability\left(\Lambda\left(\left\{\max\left\{ \frac{1}{\wb{p}_{t,e,j}^{\max}} ; \frac{z_{t+1,e,j}}{\wb{p}_{t+1,e}}\right\}\right\}_{e,j}\right) \leq a \condbar \Fp_{\{t,m,N\}}\right)\probability\left(\Fp_{\{t,m,N\}}\right).
\end{align*}
For a fixed $\Fp_{\{t,m,N\}}$, denote with $A_{e,j}$ the random variable
$\max\left\{\frac{1}{\wb{p}_{t,e,j}^{\max}}; \frac{z_{t+1,e,j}}{\wb{p}_{t+1,e}}\right\}$,
and with $A$ the set of random variables $\left\{A_{e,j}\right\}_{e,j}$ for all $e$ and $j$.
Similarly, we define $B_{e,j}$ and $B$ for the random variable
$\max\left\{\frac{1}{\wb{p}_{t,e,j}^{\max}}; \frac{z_{t,e,j}}{w_{t,e,j}}\right\}$.
Note that, given $\Fp_{\{t,m,N\}}$, $\frac{1}{\wb{p}_{t,e,j}^{\max}}$ is a constant
and $\max\left\{\frac{1}{\wb{p}_{t,e,j}^{\max}}; x\right\}$ is a monotonically
increasing function in $x$. Therefore, given $\Fp_{\{t,m,N\}}$,
$\frac{z_{t,e,j}}{w_{t,e,j}}$ stochastically dominates
$\frac{z_{t+1,e,j}}{\wb{p}_{t+1,e}}$ and
$B_{e,j}$ dominates $A_{e,j}$, since stochastic dominance is preserved by monotone functions
\cite{levy2015stochastic}.
We have
\begin{align*}
&\probability\left(\Lambda\left(\left\{\max\left\{ \frac{1}{\wb{p}_{t,e,j}^{\max}} ; \frac{z_{t+1,e,j}}{\wb{p}_{t+1,e}}\right\}\right\}_{e,j}\right) \leq a \condbar \Fp_{\{t,m,N\}}\right)\\
&= \probability\left(\Lambda\left(A\right) \leq a \condbar \Fp_{\{t,m,N\}}\right)
=\expectedvalue_{A}\left[\indfunc\left\{ \Lambda\left(A\right) \leq a \right \}\condbar \Fp_{\{t,m,N\}}\right]\\
&=\expectedvalue_{A\setminus A_{k,l}}\left[\expectedvalue_{A_{k,l}}\left[\indfunc\left\{\Lambda\left(\left\{A_{e,j}\right\}_{e,j}\right) \leq a \right\}\condbar \Fp_{\{t,m,N\}} \cap A \setminus A_{k,l}\right]\condbar \Fp_{\{t,m,N\}}\right]\\
&\stackrel{(a)}{\geq} \expectedvalue_{A\setminus A_{k,l}}\left[\expectedvalue_{B_{k,l}}\left[\indfunc\left\{\Lambda\left(\left\{A_{e,j}\right\}_{(e,j) \neq (k,l)}, B_{k,l}\right) \leq a \right\}\condbar \Fp_{\{t,m,N\}} \cap A \setminus A_{k,l}\right]\condbar \Fp_{\{t,m,N\}}\right]\\
&\stackrel{(b)}{=} \expectedvalue_{B_{k,l}}\left[\expectedvalue_{A \setminus A_{k,l}}\left[\indfunc\left\{\Lambda\left(\left\{A_{e,j}\right\}_{(e,j) \neq (k,l)}, B_{k,l}\right) \leq a \right\}\condbar \Fp_{\{t,m,N\}}\right]\condbar \Fp_{\{t,m,N\}}\right]\\
&\stackrel{(c)}{\geq} \expectedvalue_{B}\left[\indfunc\left\{\Lambda\left(\left\{B_{e,j}\right\}_{e,j}\right) \leq a \right\}\condbar \Fp_{\{t,m,N\}}\right]\\
&=\probability\left(\Lambda\left(\left\{\max\left\{  \frac{1}{\wb{p}_{t,e,j}^{\max}}; \frac{z_{t,e,j}}{w_{t,e,j}}\right\}\right\}_{e,j}\right) \leq a \condbar \Fp_{\{t,m,N\}}\right)
\end{align*}
where inequality (a) follows from the stochastic dominance and the fact that $\Lambda$ is monotonically increasing, equality (b) comes from the independence of $B_{e,j}$ from all $A$ given $\Fp_{\{t,m,N\}}$, and
inequality (c) is obtained by repeatedly applying (a) and (b) to replace all $A_{e,j}$ variables with $B_{e,j}$.
We can further iterate this inequalities (similarly to Step 4) to obtain the desired result
\begin{align*}
\probability(\normsmall{\:W_{\{t,m,N\}}} \geq \sigma^2)
&\leq \probability\left(\lambda_{\max}\left(\frac{\alpha^2(n-1)}{N^2}\sum_{j=1}^{N}\sum_{e=1}^m\left(\max_{s=1 \dots  t}\left\{\frac{z_{s-1,e,j}}{\wb{p}_{s-1,e}}\right\}\right)\:v_{e}\:v_{e}^\transp\right) \geq \sigma^2\right)\\
&\leq \probability\left(\lambda_{\max}\left(\frac{\alpha^2(n-1)}{N^2}\sum_{j=1}^{N}\sum_{e=1}^m\frac{1}{w_{0,e,j}}\:v_{e}\:v_{e}^\transp\right) \geq \sigma^2\right).
\end{align*}

\textbf{Step \arabic{cnt-lem-quad-variation}\stepcounter{cnt-lem-quad-variation} (concentration inequality).}
Since all $w_{0,e,j}$ are (unconditionally) independent from each other, we can apply the following theorem.
\begin{proposition}[Theorem~5.1.1~\cite{tropp2015an-introduction}]\label{prop:matrix-chernoff}
Consider a finite sequence $\{ \:X_k : k = 1, 2, 3, \dots \}$ whose values are independent, random, PSD Hermitian matrices with dimension $d$.  Assume that each term in the sequence is uniformly bounded in the sense that
\begin{align*}
\lambda_{\max}( \:X_k ) \leq L
\quad\text{almost surely}
\quad\text{for $k = 1, 2, 3, \dots$}.
\end{align*}
Introduce the random matrix
$\:V \eqdef \sum_{k}  \:X_k$, and
the maximum eigenvalue of its expectation
\begin{align*}
\mu_{\max} \eqdef \lambda_{\max}(\expectedvalue\left[\:V\right]) = \lambda_{\max}\left(\sum_{k} \expectedvalue\left[\:X_k\right]\right).
\end{align*}
Then, for all $h \geq 0$,
\begin{align*}
\probability\left(  \lambda_{\max}(\:V) \geq (1+h)\mu_{\max}  \right)
	&\leq d \cdot \left[\frac{e^{h}}{(1+h)^{1+h}}\right]^{\frac{\mu_{\max}}{R}}\\
	&\leq d \cdot \exp \left\{ - \frac{\mu_{\max}}{R}((h+1)\log(h+1) - h) \right\}\cdot
\end{align*}
\end{proposition}

In our case, we have
\begin{align*}
\:X_{\{e,j\}}
= \frac{\alpha^2(n-1)}{N^2}\frac{1}{w_{0,e,j}}\:v_{e}\:v_{e}^\transp
\preceq \frac{\alpha^2(n-1)}{N^2}\frac{\alpha^2}{p_{t,e}}\:v_{e}\:v_{e}^\transp
\preceq \frac{\alpha^4 (n-1)^2 a_e r_{m,e}}{a_e r_{m,e} N^2}\:I,
\end{align*}
where the first inequality follows from~\eqref{eq:distro.w} and the second
from the fact that $\:v_{e}\:v_{e}^\transp \preceq \normsmall{\:v_{e}\:v_{e}^\transp}\:I =\normsmall{\:v_{e}^\transp\:v_{e}}\:I = a_e r_{m,e} \:I$.

Therefore, we can use $L \eqdef \alpha^4(n-1)^2/N^2$ for the purpose of Proposition~\ref{prop:matrix-chernoff}. We need now to compute $\expectedvalue\left[\:X_k\right]$,
that we can use in turn to compute $\mu_{\max}$. We begin by computing the expected value of $1/w_{0,e,j}$.
Let as denote the c.d.f. of $1/w_{0,e,j}$ as
\begin{align*}
F_{1/w_{0,e,j}}(a) = \probability\left(\frac{1}{w_{0,e,j}} \leq a\right).
\end{align*}
Since $\probability\left(1/w_{0,e,j} \ge 0\right) = 1$ by~\eqref{eq:distro.w}, we have that 
\newcommand*\diff{\mathop{}\!\mathrm{d}}
\begin{align*}
&\expectedvalue\left[\frac{1}{w_{0,e,j}}\right]
= \int_{a=0}^\infty \left[1-F_{1/w_{0,e,j}}(a)\right]\diff a\\
&= \int_{a=0}^1 \left(1 - F_{1/w_{0,e,j}}(a)\right)\diff a + \int_{a=1}^{\alpha^2/p_{t,e}} \left(1 - F_{1/w_{0,e,j}}(a)\right)\diff a + \int_{a=\alpha^2/p_{t,e}}^{\infty} \left(1 - F_{1/w_{0,e,j}}(a)\right)\diff a\\
&= \int_{a=0}^1 \left(1 - 0\right)\diff a + \int_{a=1}^{\alpha^2/p_{t,e}} \left(1 - \left(1 - \frac{1}{a}\right) \right)\diff a +\int_{a=\alpha^2/p_{t,e}}^{\infty}(1-1)\diff a \\
&= \int_{a=0}^1 \diff a + \int_{a=1}^{\alpha^2/p_{t,e}} \frac{1}{a}\diff a
= 1+\log(\alpha^2/p_{t,e}),
\end{align*}
where we used the definition of the c.d.f.\,of $1/w_{0,e,j}$ in Eq.~\ref{eq:distro.w}. Thus we have
\begin{align*}
\expectedvalue\left[\:X_{\{e,j\}}\right]
= \frac{\alpha^2(n-1)}{N^2}\expectedvalue\left[\frac{1}{w_{0,e,j}}\right]\:v_{e}\:v_{e}^\transp
= \frac{\alpha^2(n-1)}{N^2}\left(1 + \log(\alpha^2/p_{t,e})\right)\:v_{e}\:v_{e}^\transp.
\end{align*}
Therefore,
\begin{align*}
&\mu_{\max} = \lambda_{\max}(\expectedvalue\left[\:V\right]) = \lambda_{\max}\Big(\sum_{\{e,j\}} \expectedvalue\left[\:X_{\{e,j\}}\right]\Big)
= \lambda_{\max}\left(\frac{\alpha^2(n-1)}{N^2}\sum_{j=1}^{N}\sum_{e=1}^m\left(1+\log(\alpha^2/p_{t,e})\right)\:v_{e}\:v_{e}^\transp\right).
\end{align*}
We now find more explicit upper and lower bounds on $\mu_{\max}$. 
We know that
\begin{align*}
p_{t,e} &\geq p_{m,e} = \frac{a_er_{m,e}}{n-1}
= \frac{a_e\:b_e^\transp \:L_\Gg^+ \:b_e^\transp}{ n-1 }
\geq \frac{a_e\lambda_{\min}(\:L_\Gg^+)\normsmall{\:b_e}^2}{ n-1 }
=\frac{2a_e}{\lambda_{\max}(\:L_\Gg)(n-1)}
\geq \frac{2a_{\min}}{a_{\max}n(n-1)},
\end{align*}
where $\lambda_{\min}(\:L_\Gg^+) = 1/\lambda_{\max}(\:L_\Gg)$ is the smallest
non-zero eigenvalue of $\:L_\Gg^+$. 
To show that $\lambda_{\max}(\:L_\Gg) \leq a_{\max}n$,
denote with $\mathcal{K}_n$ the unweighted complete graph over $n$ edges.
We use the fact that
\begin{align*}
\lambda_{\max}(\:L_\Gg) &= \max_{\normsmall{\:x} = 1} \:x^\transp \:L_\Gg \:x
=\max_{\normsmall{\:x} = 1} \frac{1}{2}\sum_{e=1}^m a_{e} (x_{e_{i}} - x_{e_j})^2\\
&\leq \max_{\normsmall{\:x} = 1} \frac{a_{\max}}{2}\sum_{e=1}^m (x_{e_{i}} - x_{e_j})^2
\leq a_{\max} \max_{\normsmall{\:x} = 1} \frac{1}{2}\sum_{e=1}^{n^2} (x_{e_{i}} - x_{e_j})^2\\
&=a_{\max}\lambda_{\max}(\:L_{\mathcal{K}_n}) \leq a_{\max}n
\end{align*}
where in the first inequality bounds summation over the weighted graph $\Gg$
with a summation over an unweighted version of $\Gg$ times $a_{\max}$,
in the second inequality we passed from a summation over the graph $\Gg$
to a summation over the full graph $\mathcal{K}_n$ and in the last
passage we bounded the largest eigenvalue of the $\:L_{\mathcal{K}_{n}}$.
Using the definition $\kappa^2 = a_{\max}/a_{\min}$ and the assumption $\alpha \leq \sqrt{\kappa n/3}$, then we obtain
\begin{align*}
\mu_{\max} & \leq \lambda_{\max}\left(\frac{\alpha^2(n-1)}{N}\sum_{e=1}^m (1 + \log(\kappa^2 n^2\alpha^2/2))\:v_{e}\:v_{e}^\transp\right)\\
&\leq \frac{ a_{\max} (1 + 2\log(\kappa n) + 2\log(\alpha))\alpha^2(n-1)}{N}\lambda_{\max}\left(\sum_{e=1}^m\:v_{e}\:v_{e}^\transp\right)\\
&\leq \frac{ 3\alpha^2(n-1)\log(\kappa n)}{N}\lambda_{\max}\left(\:P\right)
\leq \frac{3\alpha^2(n-1)\log(\kappa n)}{N}\cdot
\end{align*}
Furthermore, we have that
\begin{align*}
\mu_{\max} &\geq \lambda_{\max}\left(\frac{\alpha^2(n-1)}{N}\sum_{e=1}^m \:v_{e}\:v_{e}^\transp\right)
= \frac{\alpha^2(n-1)}{N}\lambda_{\max}\left(\:P\right) \geq \frac{\alpha^2(n-1)}{N}\cdot
\end{align*}
Therefore, selecting $h = 2$ and applying Proposition~\ref{prop:matrix-chernoff} we have
\begin{align*}
\probability&\left(\lambda_{\max}\left(\frac{\alpha^2(n-1)}{N^2}\sum_{j=1}^{N}\sum_{e=1}^m\frac{1}{w_{0,e,j}}\:v_{e}\:v_{e}^\transp\right) \geq \frac{9\alpha^2(n-1)\log(\kappa n)}{N}\right)\\
&\leq \probability\left(\lambda_{\max}\left(\frac{\alpha^2(n-1)}{N^2}\sum_{j=1}^{N}\sum_{e=1}^m\frac{1}{w_{0,e,j}}\:v_{e}\:v_{e}^\transp\right) \geq (1 + 2)\mu_{\max}\right)\\
&\leq n \cdot \exp \left\{ - \mu_{\max}\frac{N^2}{\alpha^4(n-1)^2}(3\log(3) - 2) \right\}\\
&\leq n \cdot \exp \left\{ - \frac{\alpha^2(n-1)}{N}\frac{N^2}{\alpha^4(n-1)^2}(3\log(3) - 2) \right\}\\
&\leq n \cdot \exp \left\{ - \frac{N}{\alpha^2(n-1)}\right\}\cdot
\end{align*}

\makeatletter{}
\vspace{-0.05in}
\section{Proof of Lemma \ref{lem:space-concentration} (space complexity)}
\vspace{-0.05in}
\begin{proof}[\textbf{Proof of Lemma \ref{lem:space-concentration}}]
Denote with $A$ the event $A = \left\{\forall s \in \{1, \dots, t\} :\normsmall{\wh{\:Y}_{\{s,m,N\}}} \leq \varepsilon\right\}$,
we reformulate
\begin{align*}
    &\probability\left( \sum_{j=1}^N\sum_{e=1}^t \wh{z}_{t,e,j} \geq 3N \cap \left\{\forall s \in \{1, \dots, t\} :  \normsmall{\wh{\:Y}_{\{s,m,N\}}} \leq \varepsilon\right\}\right)\\
    &= \probability\left(\sum_{j=1}^N\sum_{e=1}^t  \wh{z}_{t,e,j} \geq 3N \cap A\right)
=\probability\left(  \sum_{j=1}^{N}\sum_{e=1}^t \wh{z}_{t,e,j} \geq 3N \condbar A \right)
\probability\left(A \right)
\end{align*}
While we do know that the $\wh{z}_{t,e,j}$ are Bernoulli random variables (since
they are either 0 or 1), it is not easy to compute the success probability
of each $\wh{z}_{t,e,j}$, and in addition there could be dependencies between
$\wh{z}_{t,e,j}$ and $\wh{z}_{t,e',j'}$. Similarly to Lemma \ref{lem:prob-dominance},
we are going to find a stochastic variable to dominate $\wh{z}_{t,e,j}$.
Denoting with $u'_{s,e,j} \sim \mathcal{U}(0,1)$ a uniform random variable,
we will define $w'_{s,e,j}$ as
\begin{align*}
w'_{s,e,j} \vert \F_{\{s,e',j'\}} = w'_{s,e,j} \vert   \F_{\{s-2,m,N\}}  \eqdef \indfunc\left\{u'_{s,e,j} \leq \frac{p_{t,e}}{\wt{p}_{s-1,e}}\right\} \sim \mathcal{B}\left(\frac{p_{t,e}}{\wt{p}_{s-1,e}}\right)
\end{align*}
for any
$e'$ and $j'$ such that $\{s,1,1\} \leq \{s,e',j'\} <\{s,e,j\}$.
Note that $w'_{s,e,j}$, unlike $\wh{z}_{s,e,j}$, does not have a recursive
definition, and its only dependence on any other variable comes from
$\wt{p}_{s-1,e}$.
First, we peel off the last step
{\small
\begin{align*}
    \probability&\left( \sum_{j=1}^{N} \sum_{e=1}^t \wh{z}_{t,e,j} \geq gN \condbar A\right)\\
    &=\sum_{\Fp_{\{t-1,m,N\}}}\probability\left( \sum_{j=1}^{N} \sum_{e=1}^t \indfunc\left\{u_{t,e,j} \leq \frac{\wt{p}_{t,e}}{\wt{p}_{t-1,e}}\right\}\wh{z}_{t-1,e,j} \geq gN \condbar \Fp_{\{t-1,m,N\}} \cap A\right)\probability\left(\Fp_{\{t-1,m,N\}} \condbar A\right)\\
    &\leq \sum_{\Fp_{\{t-1,m,N\}}}\probability\left( \sum_{j=1}^{N} \sum_{e=1}^t \indfunc\left\{u'_{t,e,j} \leq \frac{p_{t,e}}{\wt{p}_{t-1,e}}\right\}\wh{z}_{t-1,e,j} \geq gN \condbar \Fp_{\{t-1,m,N\}} \cap A\right)\probability\left(\Fp_{\{t-1,m,N\}}\condbar A\right)\\
    &= \probability\left( \sum_{j=1}^{N} \sum_{e=1}^t w'_{t,e,j}\wh{z}_{t-1,e,j} \geq gN \condbar A\right),
\end{align*}
}
where we used the fact that conditioned on $A$, Prop.~\ref{prop:alpha.good} holds and
guarantees that $\wt{p}_{t,e}$ is $\alpha$-good, and therefore 
$\wt{p}_{t,e} \leq p_{t,e}$.
Plugging this in the previous bound,
\begin{align*}
\probability&\left( \sum_{j=1}^{N} \sum_{e=1}^t \wh{z}_{t,e,j} \geq gN \condbar A \right)
\probability\left(A \right)
\leq\probability\left( \sum_{j=1}^{N} \sum_{e=1}^t w'_{t,e,j}\wh{z}_{t-1,e,j} \geq gN \condbar A\right)
\probability\left(A \right)\\
&=\probability\left( \sum_{j=1}^{N} \sum_{e=1}^t w'_{t,e,j}\wh{z}_{t-1,e,j} \geq gN \cap A\right)
\leq\probability\left( \sum_{j=1}^{N} \sum_{e=1}^t w'_{t,e,j}\wh{z}_{t-1,e,j} \geq gN \right).
\end{align*}
We now proceed by peeling off layers from the end of the chain one by one. We show how to move from an iteration $s\leq t$ to $s-1$.
{\footnotesize
\begin{align*}
    \probability&\left( \sum_{j=1}^{N} \sum_{e=1}^t w'_{s,e,j}\wh{z}_{s-1,e,j} \geq gN \right)\\
    &= \expectedvalue_{\Fp_{\{s-2,m,N\}}}\left[\probability\left( \sum_{j=1}^{N} \sum_{e=1}^t \indfunc\left\{u'_{s,e,j} \leq \frac{p_{t,e}}{\wt{p}_{s-1,e}}\right\}\wh{z}_{s-1,e,j} \geq gN \condbar \Fp_{\{s-2,m,N\}} \right)\right]\\
    &= \expectedvalue_{\Fp_{\{s-2,m,N\}}}\left[\probability\left( \sum_{j=1}^{N} \sum_{e=1}^t \indfunc\left\{u'_{s,e,j} \leq \frac{p_{t,e}}{\wt{p}_{s-1,e}}\right\}\indfunc\left\{u_{s-1,e,j} \leq \frac{\wt{p}_{s-1,e}}{\wt{p}_{s-2,e}}\right\}\wh{z}_{s-2,e,j} \geq gN \condbar \Fp_{\{s-2,m,N\}} \right)\right]\\
    &= \expectedvalue_{\Fp_{\{s-2,m,N\}}}\left[\probability\left( \sum_{j=1}^{N} \sum_{e=1}^t \indfunc\left\{u'_{s-1,e,j} \leq \frac{p_{t,e}}{\wt{p}_{s-2,e}}\right\}\wh{z}_{s-2,e,j} \geq gN \condbar \Fp_{\{s-2,m,N\}} \right)\right]\\
    &= \probability\left( \sum_{j=1}^{N} \sum_{e=1}^t w'_{s-1,e,j}\wh{z}_{s-2,e,j} \geq gN \right)
\end{align*}}
Applying this repeatedly from $s=t$ to $s=2$ we have,
\begin{align*}
&\probability\left( \sum_{j=1}^{N} \sum_{e=1}^t \wh{z}_{t,e,j} \geq gN \cap \left\{\forall \; s \in \{1, \dots, t\} :\normsmall{\wh{\:Y}_{\{s,m,N\}}} \leq \varepsilon\right\} \right)\\
&\leq\probability\left( \sum_{j=1}^{N} \sum_{e=1}^t w'_{t,e,j}\wh{z}_{t-1,e,j} \geq gN \right)
=\probability\left( \sum_{j=1}^{N} \sum_{e=1}^t w'_{1,e,j}\wh{z}_{0,e,j} \geq gN \right)
=\probability\left( \sum_{j=1}^{N} \sum_{e=1}^t w'_{1,e,j} \geq gN \right). 
\end{align*}
 Now, all the $w'_{1,e,j}$ are independent Bernoulli random variables,
 and we can bound their sum with a Hoeffding-like bound using Markov inequality,
\begin{align*}
 \probability&\left( \sum_{j=1}^{N} \sum_{e=1}^t w'_{1,e,j} \geq gN \right) = \inf_{\theta > 0}\probability\left( e^{\sum_{j=1}^{N} \sum_{e=1}^t \theta w'_{1,e,j}} \geq e^{\theta g N} \right)\\
    &\leq \inf_{\theta > 0} \frac{\expectedvalue \left[ e^{\sum_{j=1}^{N} \sum_{e=1}^t \theta w'_{1,e,j}}\right]}{e^{\theta gN}}
    = \inf_{\theta > 0} \frac{\expectedvalue \left[ \prod_{j=1}^{N} \prod_{e=1}^t e^{ \theta w'_{1,e,j}}\right]}{e^{\theta gN}}
    = \inf_{\theta > 0} \frac{\prod_{j=1}^{N} \prod_{e=1}^t \expectedvalue \left[ e^{ \theta w'_{1,e,j}}\right]}{e^{\theta gN}}\\
    &=\inf_{\theta > 0} \frac{\prod_{j=1}^{N} \prod_{e=1}^t (p_{t,e} e^\theta + (1-p_{t,e}))}{e^{\theta gN}}
    = \inf_{\theta > 0} \frac{\prod_{j=1}^{N} \prod_{e=1}^t (1+p_{t,e}( e^\theta -1))}{e^{\theta gN}}\\
    &\leq \inf_{\theta > 0} \frac{\prod_{j=1}^{N} \prod_{e=1}^t e^{p_{t,e}( e^\theta -1)}}{e^{\theta gN}}
    \leq \inf_{\theta > 0} \frac{e^{N( e^\theta -1)}}{e^{\theta gN}}
    = \inf_{\theta > 0} e^{(N e^\theta -N - \theta gN)},
\end{align*}
where we use the fact that $1 + x \leq e^x$ and by definition $w'_{1,e,j} \sim \mathcal{B}(p_{e,t})$ and $\sum_{e=1}^t p_{t,e} = 1$.

The choice of $\theta$ minimizing the previous expression is obtained as
\begin{align*}
    \frac{d}{d\theta}e^{\left( N e^\theta -N - \theta gN\right)}
    = e^{\left( Ne^\theta -N - \theta gN\right)}\left(Ne^\theta - g N\right) = 0,
\end{align*}
and thus $\theta = \log (g)$. Finally,
\begin{align*}
    \probability\left( \sum_{j=1}^{N} \sum_{e=1}^t \wh{z}_{m,e,j} \geq gN \right)
    \leq \inf_\theta \exp\left\{ N(e^\theta - 1 - \theta g)\right\}
    = \exp\left\{ N\left(g  - 1 -  g \log (g)\right)\right\}
\end{align*}
choosing $g = 3$ and plugging in the definition of $N$ from Algorithm \ref{alg:kl_sequence},
\begin{align*}
\probability&\left( \sum_{j=1}^{N} \sum_{e=1}^t \wh{z}_{t,e,j} \geq 3N \cap \left\{\forall \; s \in \{1, \dots, t\} :  \normsmall{\wh{\:Y}_{\{s,m,N\}}} \leq \varepsilon\right\}\right)\\
&\leq \exp\left\{ -40\alpha^2 n \log^2 (3m/\delta) / \vareps^2\right\} \leq \exp\left\{-\log(2m/\delta)\right\} = \frac{\delta}{2m}\cdot
\end{align*}
\end{proof}

\bibliographystyle{abbrvnat}

\newpage
\clearpage

\end{document}